# On Data-Independent Properties for Density-Based Dissimilarity Measures in Hybrid Clustering

**Kajsa Møllersen[1], Subhra S. Dhar[2], Fred Godtliebsen[3]**

[1]Norwegian Centre for Integrated Care and Telemedicine, University Hospital of North Norway, Tromsø, Norway
[2]Department of Mathematics and Statistics, IIT Kanpur, Uttar Pradesh, India
[3]Department of Mathematics and Statistics, UiT The Arctic University of Norway, Tromsø, Norway
Email: kajsa.mollersen@gmail.com, subhra@iitk.ac.in, fred.godtliebsen@uit.no

## Abstract

Hybrid clustering combines partitional and hierarchical clustering for computational effectiveness and versatility in cluster shape. In such clustering, a dissimilarity measure plays a crucial role in the hierarchical merging. The dissimilarity measure has great impact on the final clustering, and data-independent properties are needed to choose the right dissimilarity measure for the problem at hand. Properties for distance-based dissimilarity measures have been studied for decades, but properties for density-based dissimilarity measures have so far received little attention. Here, we propose six data-independent properties to evaluate density-based dissimilarity measures associated with hybrid clustering, regarding equality, orthogonality, symmetry, outlier and noise observations, and light-tailed models for heavy-tailed clusters. The significance of the properties is investigated, and we study some well-known dissimilarity measures based on Shannon entropy, misclassification rate, Bhattacharyya distance and Kullback-Leibler divergence with respect to the proposed properties. As none of them satisfy all the proposed properties, we introduce a new dissimilarity measure based on the Kullback-Leibler information and show that it satisfies all proposed properties. The effect of the proposed properties is also illustrated on several real and simulated data sets.

## 1. Introduction

Clustering or cluster analysis is a well-known unsupervised statistical learning tech-nique, which describes the formation of groups in a data set based on the relation between the observations. The common understanding of a cluster is that the observations within the cluster are more closely related to each other than to the observations in another cluster. The definition of closely related observations depends both on the problem at hand and the user's understanding of the cluster concept, and hence there exists a wide range of clustering techniques. **Figure 1** shows one way of categorising clustering techniques, which will be helpful for the understanding of hybrid clustering and dissimilarity measures. At the top level, clustering techniques are divided into partitional and hierarchical clustering. Partitional clustering provides a single partition of the data, whereas hierarchical clustering provides a nested series of partitions. At each step in a hierarchical clustering, either two subclusters are merged (agglomerative clustering) or one subcluster is divided into two subclusters (divisive clustering). A dissimilarity measure defines which two subclusters will be merged at each step in agglomerative clustering, and might be based on a probability density function or a proper distance function. Several other categorisations of clustering techniques exist; for a comprehensive overview, we refer the readers to [1]-[3].

*Hybrid clustering* combines partitional and hierarchical clustering to overcome the shortcomings of each separate technique. Model-based partitional clusterings, like Gaussian mixture model (GMM) and $k$-means [4] [5], impose shape restrictions on the clusters [6], which are a drawback if the model does not fit the data well. Model-free partitional clusterings are more versatile, but for high-dimensional data, sparsity makes low- and high-density regions difficult to distinguish. On the other hand, the hierarchical clustering technique does not impose restrictions on the cluster shape, but it has extensive computational complexity. Hybrid clustering takes advantage of the versatility of hierarchical clustering, but avoids the computational costs, and can be described as follows: Initially, a partitional clustering defines $K$ subclusters. These are then merged hierarchically into $C$ final clusters. Hybrid clustering and its related issues have been studied for decades; see e.g. [6]-[15].

To illustrate how different clustering techniques give us different outcomes, let us consider the well-known Old Faithful Geyser data, which is available in *R*. The data set consists of the length of 272 eruptions, and eruption length versus previous eruption length is shown in **Figure 2(a)**. The observations are clearly clustered, and the objective of a clustering technique is to identify the clusters. The actual number of clusters is unknown, and it has served as an illustrative example for a number of techniques. It illustrates the clustering concept through a set of observations where some are more closely related than others, but also the inherit subjectivity since not all observations have a clear cluster membership, and the number of clusters depends on the user's subjective understanding. Clustering includes both assigning the observations to a specific cluster and estimating the number of clusters. Depending on the observer's subjective understanding, there are either three or four clusters in Figure 2(a), and different cluster number estimation techniques give different results. For a discussion on three versus four clusters for this particular data set, see [16]. Figure 2(b) shows an example of a clustering result where the Bayesian information criterion (BIC) was used to estimate the number of clusters, and the observations have been clustered using GMM and labelled according to the maximum a posteriori probability (MAP). Alternatively, the gap statistic (see Section 2.2) can be used to estimate the number of clusters, which resulted in three clusters. The partitional clustering in Figure 2(b) can then be regarded as the initial step in a hybrid clustering, where BIC was used to estimate the number of *subclusters* $K$. The next and final step will then be to merge two of the subclusters to reach the number of final clusters, $C = 3$. It turns out that the choice of dissimilarity measure has significant impact on the final result of hybrid clustering. The well-known distance-based dissimilarity measures, single-link, complete-link and average-link merge blue-green coloured, blue-red coloured and blue-green coloured observations, respectively. On the other hand, the measures *SE* (see Section 3.3) and *Bhat* (see Section 3.3), both density-based, merge blue-green coloured and blue-red coloured observations, respectively. There is no definition of "true" clusters, but some results are more agreeable with the cluster concept. The results when using single-link, average-link or *SE* as dissimilarity measure, have, to our knowledge, not been proposed for this well-known data-set.

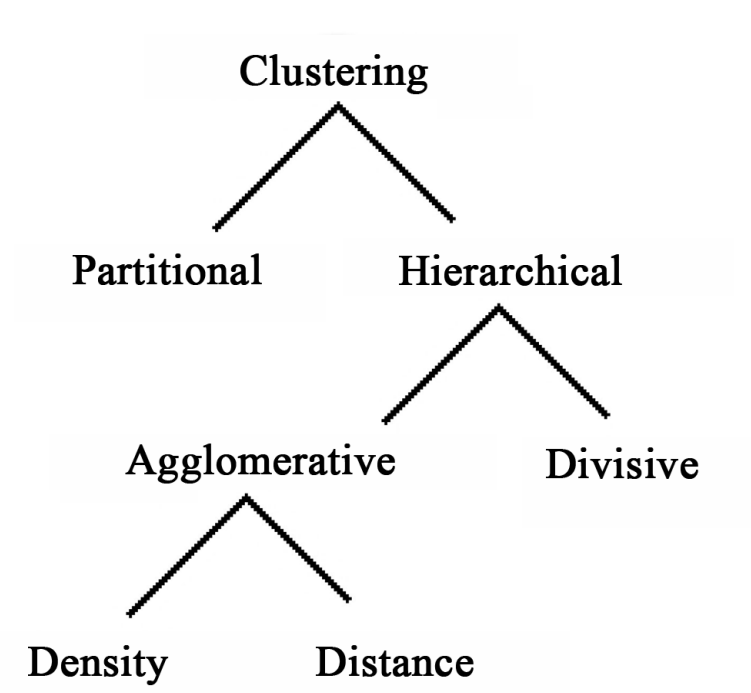

**Figure 1.** A partial taxonomy of clustering techniques.

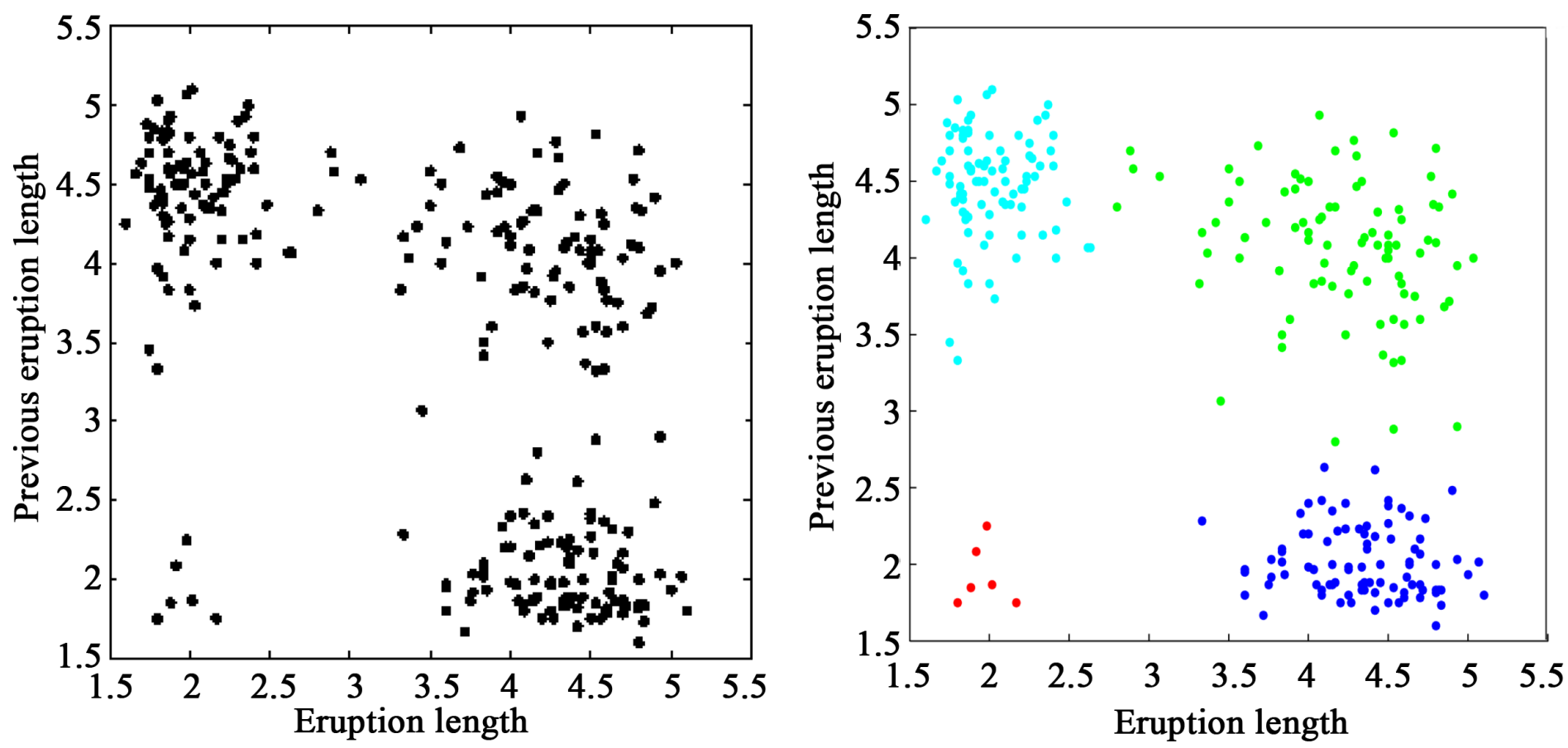

(a) (b)

**Figure 2.** Eruption length versus previous eruption length of the Old Faithful Geyser. (a) The gap-statistic estimates three clusters, whereas ICL estimates four (see Section 2.2); (b) Data labelled according to MAP for a four-component GMM fitted to the data. The number of components was estimated by BIC (see Section 2.1).

In the above example, the choice of dissimilarity measure can be done based on whether the result agrees with the user's own perception. However, when the dimension is larger than three, the user cannot check the clustering result by visual inspection. If the user already has a clear opinion on how the data should be clustered, a clustering technique is not needed. For other situations, when clustering is used to explore the data, to discover patterns or as a pre-processing for other tasks, the clustering result must be interpreted in view of the method used. The result can be interpreted in view of how the chosen dissimilarity measure behaves under various circumstances, or *vice-versa*; the dissimilarity measure can be chosen based on pre-defined requests. Since the same dataset can lead to very different clustering results, as illustrated in the above example, the clustering result itself says very little about the data unless some analysis of the method is included.

In the present work, we have investigated one crucial aspect of clustering; the choice of dissimilarity measure. The investigation is restricted to density-based measures, and is exemplified on the partitional-hierarchical hybrid clustering technique. At first, we have identified a set of properties that are relevant for density-based dissimilarity measures in the hybrid clustering context (see Section 3.2). The properties have been identified by observing known dissimilarity measures' behaviour under circumstances that are likely to occur in the hybrid clustering context. Further, we have analysed a set of known dissimilarity measures in view of these properties, and none of them fulfil all. Furthermore, we propose a new density-based dissimilarity measure that fulfils all the proposed properties. The new measure is constructed to be robust to background noise, a property that none of the other dissimilarity measures fulfil.

The performance of a dissimilarity measure on a specific data set does not imply that it will perform equally good or bad on another data set. Data-independent properties to evaluate dissimilarity measures enable us to choose a dissimilarity measure based on the user's understanding of the cluster concept for a given problem without *extra* knowledge about the data set. For instance, whether a small subcluster, such as the red observations in **Figure 2(b)**, should be considered a proper cluster or the effect of noise observations can be known beforehand, regardless of the actual data. The preferred concepts may differ, and data-independent properties will serve to categorise the dissimilarity measures. Properties for distance-based dissimilarity measures have been discussed for decades [2] [3] [17]-[20]. To our best knowledge, there has been no attempt in investigating data-independent properties for density-based dissimilarity measures, except in the context of image analysis [21]. However, many density-based dissimilarity measures originate from divergence functions, which possess certain properties (see Section 3.1). In this article, we propose additional properties regarding clustering-specific problems related to small outlying subclusters, low-density subclusters associated with background noise, and light-tailed models for heavy-tailed data (see Section 3.2). We would like to point out that this is not an attempt to describe the best dissimilarity measure, but rather to provide a tool for evaluation and categorisation depending upon prior knowledge on the data and the user's understanding regarding the clustering concept. In addition, several density-based dissimilarity measures have been proposed for hybrid clustering, and since none of them satisfy all the proposed properties, we propose a new one that does.

The rest of the article is organised as follows. Hybrid clustering is discussed in Section 2. Section 3 introduces the *f*-divergences and proposes a set of properties for dissimilarity measures. Section 3.3 investigates well-known dissimilarity measures and proposes a new dissimilarity measure. Section 4 illustrates the dissimilarity measures' impact on the final clustering based on different simulated and real data. Some concluding remarks are mentioned in Section 5. The appendices provide the necessary technical details.

## 2. Hybrid Clustering Techniques

As discussed in the Introduction, hybrid clustering seeks to take advantage of the versatility of hierarchical clustering, but at the same time avoid the computational complexity. Hybrid clustering can be summarised in three steps:

1) Partition the data into $K$ subclusters by partitional clustering.

2) Hierarchically merge the two subclusters with lowest dissimilarity.

3) Repeat the merging until some stopping rule applies.

Let $K$ denote the number of initial partitions, $N$ denote the number of observations, and $C$ denote the number of final clusters. When $K$ is close to $N$, the shape restrictions of the initial partitioning technique does not have significant impact on the final result. On the other hand, when $K$ is close to $C$, the computational complexity will be optimised, and hence, there is a trade-off between versatility and computational complexity. For a more thorough discussion on the computational complexity of partitional, hierarchical and hybrid clustering, see [15]. The subsequent subsections discuss each of the three steps.

### 2.1. Partitional Clustering

Various techniques have been used for initial partitioning, among them the well-known *k*-means [6] [8]-[10], and GMM [7] [8] [22]. For other examples, see [11]-[13] [23]. A brief description of the GMM is as follows.

Any continuous distribution $f(\boldsymbol{x})$ can be approximated with arbitrary accuracy by a mixture of Gaussian distributions [24] [25]:

$$f(\boldsymbol{x}) \approx f_K(\boldsymbol{x}) = \sum_{i=1}^{K} \pi_i \phi(\boldsymbol{x} \mid \boldsymbol{\mu}_i, \Sigma_i), \quad \text{where } \pi_i > 0, \sum_{i=1}^{K} \pi_i = 1.$$

Here, $\pi_i$ is the a priori probability associated with the $i$-th class, and $\boldsymbol{\mu}_i$ and $\Sigma_i$ are the location and the scale parameters of the Gaussian distribution of the $i$-th component. When the underlying distribution is unknown, the GMM is fitted to the observed values, most commonly by the EM algorithm [26], and the accuracy will depend on how the observations were generated from the parent distribution. The number of components must be estimated separately, with a trade-off between good fit and model complexity. A common procedure is to fit several GMMs, having $K_{\min},\cdots,K_{\max}$ components, and then use a criterion like BIC [27] or the Akaike information criterion (AIC) [28] to select the number of components. For other criteria and a discussion, the reader is referred to [29], p. 175.

For any choice of dissimilarity measure, the final clustering depends on the partitional clustering technique through the initial subclusters' *purity*. In a totally pure subcluster, all observations belong to the same "true" cluster, but frequently one or more subclusters will contain observations from at least two different "true" clusters. The final clustering will therefore rely on the purity of the partitional clustering. We illustrate this on the Yeast data from the UCI machine learning repository [30]. Franczak *et al.* chose three variables (McGeoch's method for signal sequence recognition (mcg), the score of the ALOM membrane spanning region prediction program (alm), and the score of discriminant analysis of the amino acid content of vacuolar and extracellular proteins (vac)), and two classes (localisation sites CYT (cytosolic or cytoskeletal) and ME3 (membrane protein, no N-terminal signal)) to illustrate Laplace clustering [31]. **Figure 3(a)** shows the observations labelled according to class.

We use GMM and BIC for partitional clustering, and the observations are assigned to a subcluster according to MAP. The result can be seen in **Figure 3(b)** where the subcluster shown as yellow crosses contains observations from both classes. **Figure 3(c)** shows the partitional clustering when using Student's *t* mixture distribution instead of GMM, and we see that the subclusters are purer. To be able to separate the impact of the dissimilarity measure from that of the partitional clustering technique (see Section 4) we define *minimum misclassification rate* as the minimum proportion of observations assigned to the wrong cluster when all possible final clusterings are considered. The minimum misclassification rates in this example are 7.8% for GMM and 3.4% for Student's *t*. The superior performance of Student's *t* in this example was expected, since Student's *t* mixture distributions are more robust to heavy tails than GMMs [32]. In general, the best partitional clustering technique will depend on the problem at hand.

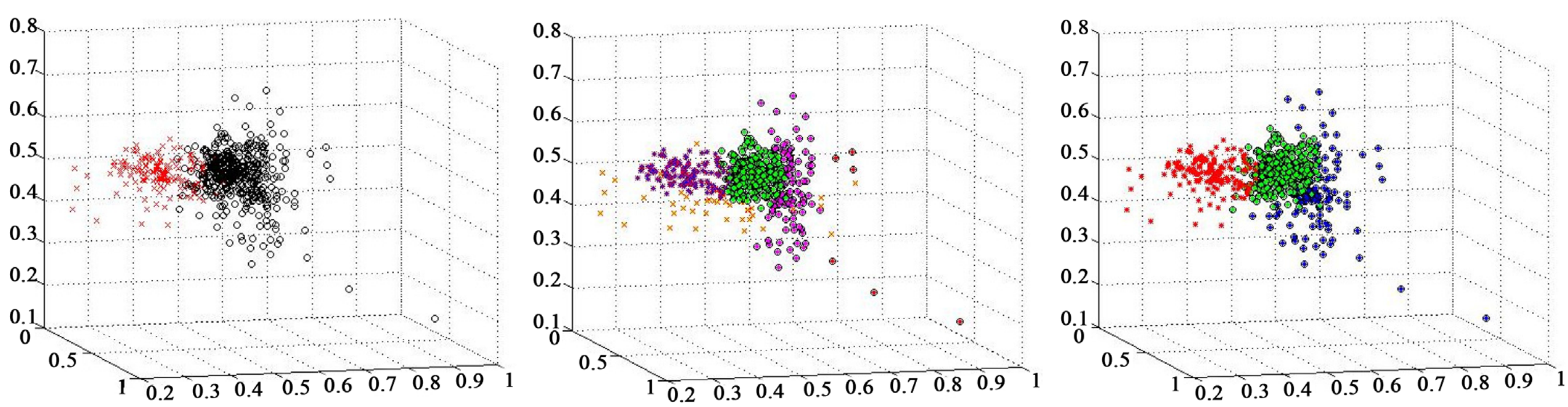

(a) (b) (c)

**Figure 3.** Yeast data. (a) The true labelling; (b) Five-component GMM fit; and (c) Three-component Student's *t* mixture fit.

## 2.2. Dissimilarity Measure and Stopping Rule

Dissimilarity measures for hybrid clustering can be based on distance functions (for example, [12]) or density functions (for example, [8]). The dissimilarity measure will be involved in each step in the hierarchical merging, and therefore, the choice of dissimilarity measure might have significant impact on the final outcome, as discussed in the Introduction. Density-based dissimilarity measures are functions of the subclusters' density and size (also referred to as weight), estimated from the data. If the partitional clustering is done by mixture distribution fitting, the components' probability density function (pdf) and a priori probability of the class represent the subclusters' density and weight, and the dissimilarity measure is a function of the observations indirectly through the estimated mixture distribution.

The number of clusters can be estimated from the data using, for example, the gap statistic [33] or integrated completed likelihood (ICL) [34]. For demonstration, an example is shown in **Figure 2(a)**. The dissimilarity measure can define the stopping rule by setting a threshold value at which the merging process will stop. This threshold can be pre-defined by the user, or defined by analysing the dissimilarity as a function of the number of subclusters, $K, K-1, \cdots, 1$ [33], see **Figure 11**. The number of clusters can also be chosen by the user based on prior knowledge about the data. To gain insight about different stopping rules, the reader may read [19].

# 3. Data-Independent Properties for Density-Based Dissimilarity Measures

The divergence between two probability distributions essentially measures the *distance* between them, and hence there is a close relationship between density-based dissimilarity measures and divergences. There have been many attempts to define the distance between two probability distributions, and the concept is context dependent and subjective like clustering. Note that many divergences do not satisfy symmetry or triangular inequality, and are therefore not proper distances. Ali and Silvey [35] defined four properties for divergences, and through them defined the class of *f*-divergences. The *f*-divergences are of special interest due to their properties and since they have been used directly or in modified form as dissimilarity measures for hybrid clustering [15] [21] [36]. Many well-known distance functions or information criteria are *f*-divergences, for example, variational distance, Hellinger/Kolmogorov distance, Kullback-Leibler information, Kullback-Leibler/Jeffrey divergence, Chernoff distance, Bhattacharyya distance, Matusita distance [35] [37]-[39], and therefore automatically satisfy the properties.

Dissimilarity measures used as merging criteria attempt to measure the distance between two subclusters with respect to which two subclusters are most appropriate to merge. We therefore introduce properties specifically for dissimilarity measures which offer the user a data-independent tool for choosing, evaluating or modifying dissimilarity measures.

### 3.1. Properties for *f*-Divergences

The *f*-divergences, introduced independently in [38] and in [35], are defined as follows [39].

**Definition:** *Let* $\boldsymbol{X}$ *be a random vector defined on a subset of* $\mathbb{R}^d$, *and* $p_k(.)$ *be the probability density function of the probability distribution function* $P_k(.)$. *For any convex function f having* $f(1)=0$ *and* $0f(0/0)=0$

$$d(P_k,P_l)=\int_{\mathcal{X}} p_l(\boldsymbol{x})\, f\left(\frac{p_k(\boldsymbol{x})}{p_l(\boldsymbol{x})}\right)\mathrm{d}\boldsymbol{x},$$

is an *f*-divergence. Here $\mathcal{X}$ is the support of the probability distribution function induced by the random vector $\boldsymbol{X}$.

The *f*-divergences satisfy, by construction, the following properties:

1) An *f*-divergence $d(P_k,P_l)$ is defined for all pairs $P_k,P_l$ on the same sample space.

2) For any measurable transformation $t(.)$, $d(P_k,P_l)\geq d(P_k t^{-1},P_l t^{-1})$.

3) The *f*-divergence $d(P_k,P_l)$ takes its minimum when $P_k=P_l$ and maximum when $P_k$ and $P_l$ are orthogonal, denoted by $P_k\perp P_l$.

4) If $\{p_\theta;\theta\in(a,b)\}$ is a family of densities on $\mathbb{R}$ with monotone likelihood ratio, then for $a<\theta_1<\theta_2<\theta_3<b$, we have $d(P_{\theta_1},P_{\theta_2})\leq d(P_{\theta_1},P_{\theta_3})$.

We will refer to them as *f*-properties. For more rigorous definitions, details, explanations, examples and discussions, see [35]. The first property simply ensures validity of the divergence. The second property, which is extensively studied in [38], implies that a measurable transformation will not increase the divergence between two probability distributions. At first, this property might seem counterintuitive in a clustering context, since transformations are often used to increase discernability between clusters. However, these transformations are applied on the observations *before* density estimation, and hence, the increased discernability will be reflected by the new density estimates. The third property stating minimum value at equality and maximum value at orthogonality follows immediately from the concept of a divergence being a measure of the distance between two probability distributions. A definition of orthogonality between distributions can be found in [40] (see also Appendix A). The fourth property ensures that when $P_{\theta_3}$ is *further away* from $P_{\theta_1}$ than $P_{\theta_2}$ is, the divergence between $P_{\theta_1}$ and $P_{\theta_3}$ will be larger than between $P_{\theta_1}$ and $P_{\theta_2}$, which also follows from the divergence being a concept of distance. There has been extensive research on *f*-divergences, and they have other properties not listed here. We have presented these four properties for the reader to gain insight about *f*-divergences since they are popular as fundaments for dissimilarity measures.

### 3.2. Properties for Dissimilarity Measures for Hybrid Clustering

An important difference between divergences and dissimilarity measures is that subclusters can have different weights (corresponding to the a priori probability), as opposed to probability distributions. Figure 4(a) shows an example of a Gaussian mixture distribution where all three components have equal variance. Since the Gaussian distribution has monotone likelihood ratio, it follows from the fourth *f*-property that the two components with smallest distance between the means will have the smallest *f*-divergence. Since the blue and the red components are slightly closer to each other than the red and the green components are, those are the two with the smallest *f*-divergence. This illustrative example shows that dissimilarity measures should be functions of both the subclusters' density and weight, and that dissimilarity measures should have other properties than divergences. We will continue this section with motivation for the data-independent properties presented here.

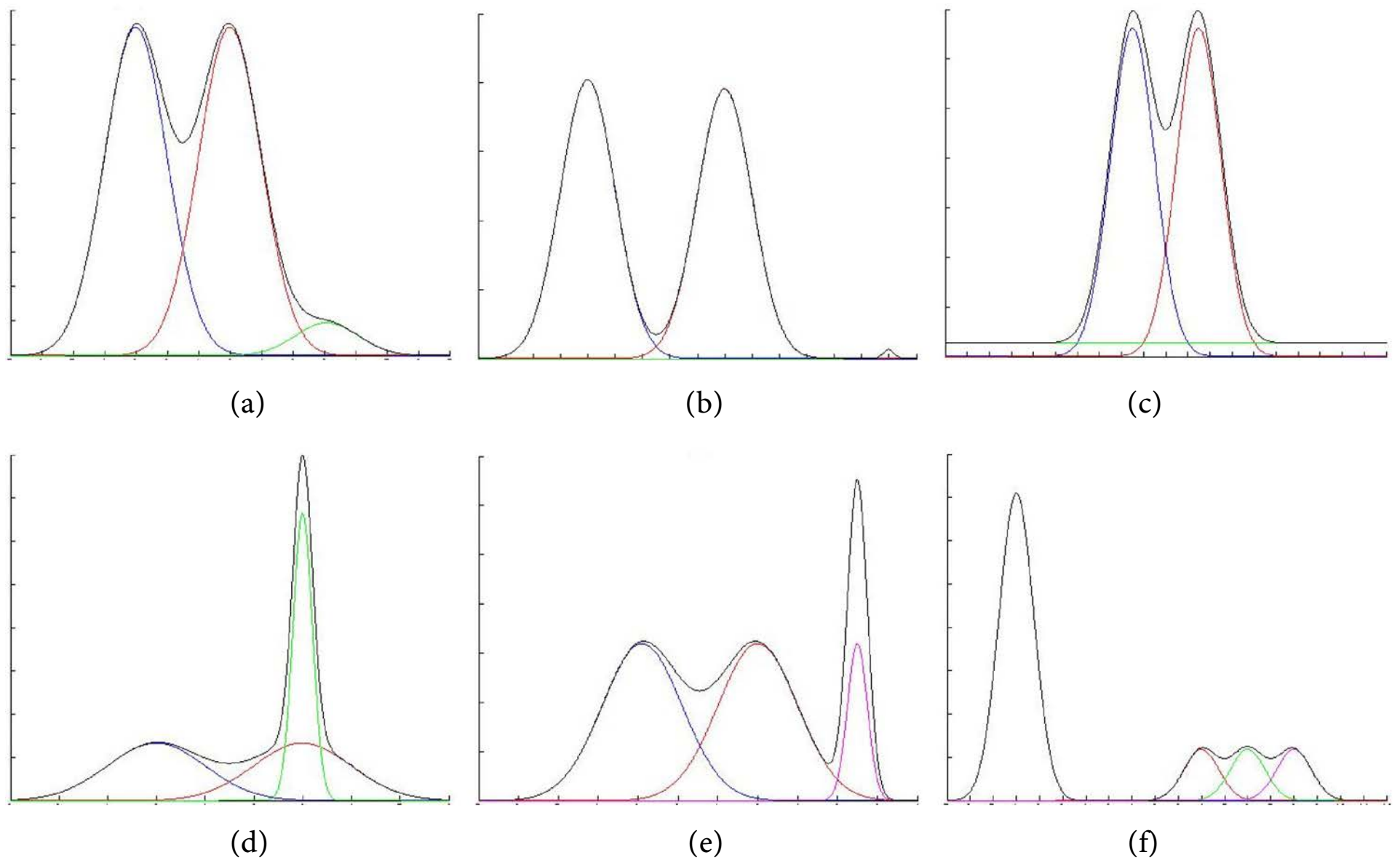

**Figure 4.** Gaussian mixtures densities (black line) and their components; coloured lines from left to right; blue, red, green, pink. (a) Equal variance, unequal a priori probabilities; (b) Small outlying component; (c) Background noise component; (d) Equal mean and unequal variance; (e) Two equal components (green hidden by pink); (f) A priori probabilities with different relative values.

**Figure 4** shows Gaussian mixture densities and their components, and illustrates different aspects of merging in the following situation: The coloured lines represent the probability densities of the current subclusters, and the next step in the hybrid clustering is to merge the two subclusters with lowest dissimilarity. These examples illustrate the demand for the properties introduced later in this section. The details are found in Appendix B, where also the results of different dissimilarity measures (see Section 3.3) are given.

Here, we illustrate that the presence of outlier and noise observations can obstruct the clustering. Because of this, we propose properties that involve robustness to outlier and noise observations. Since density-based dissimilarity measures are functions of the observations indirectly through the density and weight estimates of the subclusters, we refer to an *outlying component* as one having small weight and large distance from the other components (see **Figure 4(b)**), and *noise component* as one having large scatter and low maximum density (see **Figure 4(c)**). Robustness, in this context, refers to the concept that the presence or absence of an outlier component or noise component should not influence the final clustering significantly. This can be illustrated by the example in **Figure 4(b)**. Let the small component to the right be an outlying component, and let the stopping rule of this particular clustering be to stop at two clusters. If the outlying component were absent, the clustering would not involve merging, and the two components to the left would define the two final clusters. In the presence of the outlying component, any robust dissimilarity measure would merge the outlying component with one of the large components, as this has minimal influence on the final clustering. Similarly, for **Figure 4(c)**, the noise component should be merged with one of the other components instead of constituting a cluster of its own in the case of a two-cluster solution.

There are other ways to deal with outliers and noise than by having robust dissimilarity measures. A common approach is to remove outliers and noise observations by pre-processing of the data (for example, trimming or snipping, [41]), and for hybrid clustering, several algorithms have been proposed [8] [9] [13] [14]. The drawback of this approach is that it usually requires some user-set parameter, which in turn requires knowledge of the data. Huber and Ronchetti ([42], p. 4) discussed several other issues regarding robustness versus pre-processing. Another important issue that they addressed ([42], p. 8) is the difference between robustness and diagnostics, where the purpose of diagnostics is to detect the outliers. Clearly, if the main objective of a particular clustering procedure is diagnostics then the dissimilarity measure should *not* be robust to outliers. It is therefore important to know whether the chosen dissimilarity measure satisfies a given outlier property.

Hybrid clustering assumes that the model used for partitional clustering is different from the final clusters, else, there would be no need for a merging step. An example is the GMM for heavy-tailed clusters, as illustrated in **Figure 4(d)** where the red and the green components have the same mean, and unequal variances. Whether the red and the green component actually belong to the same cluster depends on the particular understanding of the cluster concept, and having a data-independent property for this particular situation enables the user to choose the proper dissimilarity measure.

The third *f*-property (minimum at equality and maximum at orthogonality) is fundamental for divergences, but can cause trouble for dissimilarity measures. In **Figure 4(b)**, the outlying component is nearly orthogonal to the two other components, but if the dissimilarity measure is robust to outliers, it cannot at the same time attain its maximum value. We therefore propose an additional $\pi_k = \pi_l$, where $\pi_k$ and $\pi_l$ are the weights of subcluster $k$ and $l$, to the third *f*-property to make it suitable for dissimilarity measures. We use the $\pi_i$ both to denote a priori probabilities for a mixture distribution, and the subclusters' weight, since one is an estimate of the other. **Figure 4(e)** shows a four-component GMM where the two rightmost components are identical, and therefore the green component is hidden behind the pink. The Shannon entropy dissimilarity measure, *SE*, defined in Section 3.3, does not attain its minimum value at equality, and as a result, the dissimilarity between the blue and the red component is smaller than that between the identical green and pink components.

First, we propose a symmetry property, which does not require further explanation. The other five properties are motivated by the discussions earlier in this section. Suppose that $p_k$ is the pdf associated with the *k*-th subcluster having weight $\pi_k$, where $\sum_{i=1}^{K}\pi_i = 1$. The dissimilarity between the subclusters *k* and *l* is denoted by $D(k,l)$. Let $m = \min(D)$ when $\min(D) > -\infty$, or arbitrarily small when $\min(D) = -\infty$, and let $M = \max(D)$ when $\max(D) < \infty$, or arbitrarily large when $\max(D) = \infty$.

**Symmetry:** The dissimilarity measure $D(k,l)$ is symmetric, $D(k,l) = D(l,k)$ for all *k* and *l*.

**Equality:** If $\pi_k = \pi_l$ and $p_k(\boldsymbol{x}) = p_l(\boldsymbol{x})$, then $D(k,l) = \min(D)$.

**Orthogonality:** If $\pi_k = \pi_l$ and $p_k(\boldsymbol{x}) \perp p_l(\boldsymbol{x})$, then $D(k,l) = \max(D)$.

**Outlier:** If $p_k(\boldsymbol{x})$ and $p_l(\boldsymbol{x})$ are not orthogonal, and if one of the weights is close to zero, the dissimilarity measure will be close to its minimum. That is, for every $\epsilon > 0$, there exists a $\delta > 0$ such that $\min(\pi_k, \pi_l) < \delta \Rightarrow |D(k,l) - \max(\min(D), m)| < \epsilon$.

**Noise:** If the maximum density of the large-scatter subcluster, $p_k(\boldsymbol{x})$, is close to zero, the dissimilarity measure will be close to its minimum. That is, if $\max(p_k(\boldsymbol{x})) < \max(p_l(\boldsymbol{x}))$ and $p_k(\boldsymbol{x}) \geq \min(p_l(\boldsymbol{x}))$ for all $\boldsymbol{x}$, then for every $\epsilon > 0$, there exists a $\delta > 0$ such that $\max(p_k(\boldsymbol{x})) < \delta \Rightarrow |D(k,l) - \max(\min(D), m)| < \epsilon$.

**Mode:** Let $p_k$ and $p_l$ be two Gaussian probability density functions with location parameters $\boldsymbol{\mu}_k$ and $\boldsymbol{\mu}_l$, respectively, and scatter parameters $\Sigma_k$ and $\Sigma_l$, respectively. Further, suppose that $\boldsymbol{\mu}_k = \boldsymbol{\mu}_l$ and $\Sigma_l = a\Sigma_k$, where $a \in (0,1)$ is a constant. Under such circumstances, for every $\epsilon > 0$, there exists a $\delta > 0$ such that $a < \delta \Rightarrow |D(k,l) - \min(\max(D), M)| < \epsilon$. And, for every $\epsilon^* > 0$, there exists a $\delta^* > 0$ such that $a > 1 - \delta^* \Rightarrow |D(k,l) - \max(\min(D), m)| < \epsilon^*$.

The properties do not state whether the outlying component in Figure 4(b), the noise component in Figure 4(c), or the two components with equal mean in Figure 4(d) should be merged. They state that, *at some point*, for a very small component, for a component with very low maximum density, or for two components with equal mean, the dissimilarity measure should attain its minimum value.

### 3.3. Dissimilarity Measures for Hybrid Clustering

We here present some density-based dissimilarity measures, illustrating which properties some well-known dissimilarity measures satisfy. Many dissimilarity measures are closely connected, e.g. the Shannon entropy and the Kullback-Leibler information, but they satisfy different properties. The list of dissimilarity measures is not, and cannot be, extensive, but shows a variety and includes dissimilarity measures both based on *f*-divergences and the Shannon entropy. None of the dissimilarity measures we could find in the literature satisfies all of the properties, and we therefore propose one that does; a Kullback-Leibler information based dissimilarity measure, *KLinf*. Table 1 shows which properties each dissimilarity measure satisfies. The technical details are provided in Appendix A. In the following, $p_k$ and $p_l$ are the pdf's of the probability measures $P_k$ and $P_l$, respectively, and $\pi_k$ and $\pi_l$ are the weights of the *k*-th and the *l*-th subcluster, respectively, and $w_k = \pi_k / (\pi_k + \pi_l)$ and $w_l = \pi_l / (\pi_k + \pi_l)$.

**1) Shannon entropy**

The Shannon entropy for a continuous distribution, $f(x)$, is defined as [43]

$$H(f) = -\int f(x) \log f(x) \mathrm{d}x.$$

We here present three dissimilarity measures for probability measures $P_k$ and $P_l$ that are functions of the Shannon entropy; the Shannon entropy dissimilarity measure (*SE*), the weighted Shannon entropy dissimilarity measure (*wSE*), and the Jensen-Shannon dissimilarity measure (*JS*).

$$SE(k,l) = H(\pi_k p_k + \pi_l p_l) - H(\pi_k p_k) - H(\pi_l p_l)$$

$$wSE(k,l) = (\pi_k + \pi_l) H(p_k + p_l) - \pi_k H(p_k) - \pi_l H(p_l)$$

$$JS(k,l) = H(w_k p_k + w_l p_l) - w_k H(p_k) - w_l H(p_l)$$

*SE* minimises the Shannon entropy of the resulting merging [22]. *wSE* measures the divergence between the two components and their merging, multiplied by the subclusters' weights [36]. Straightforward algebra shows that *JS* is a weighted version of the Kullback-Leibler divergence [44].

As discussed in Section 3.2, the Shannon entropy dissimilarity measure does not satisfy the Equality property, with the result seen in **Figure 4(e)**. The weighted version, *wSE*, neither satisfies the Equality nor the Orthogonality property. The Jensen-Shannon dissimilarity measure is a function of relative weights, and an example of its consequences is seen in **Figure 4(f)**, showing a large component to the left (blue) and three equally sized components to the right (red, green, and pink). Since the red component is relatively small compared to the blue one, but has the same size as the green one, the Jensen-Shannon dissimilarity between the blue and the red component is smallest.

**Table 1.** Dissimilarity measures and properties. "x" indicates satisfaction and "-" indicates violation.

| Dissimilarity measure | Property | | | | | |
|---|---|---|---|---|---|---|
| | Equality | Orthogonality | Symmetry | Outlier | Noise | Mode |
| SE | - | x | x | - | - | - |
| wSE | - | - | x | - | - | - |
| JS | x | - | x | - | - | - |
| Err | x | x | x | - | - | x |
| Bhat | x | x | x | x | - | x |
| KLdiv | x | x | x | x | - | x |
| KLinf | x | x | x | x | x | x |

**2) Probability of misclassification**

The error probability of the Bayes classifier for two classes is $\int \min\left(w_k, p_k(\boldsymbol{x}), w_l p_l(\boldsymbol{x})\right) \mathrm{d}\boldsymbol{x}$, and subtracting it from one will be an *f*-divergence [37]

$$Err(k,l) = 1 - \int \min\left(w_k p_k(\boldsymbol{x}), w_l p_l(\boldsymbol{x})\right) \mathrm{d}\boldsymbol{x}.$$

*Err* is an example of a dissimilarity measure that does not satisfy the Outlier property. As a consequence, the small outlying component in **Figure 4(b)** will not be merged with any of the other components even if its size decreases to a minimum.

**3) Bhattacharyya distance**

The Bhattacharyya coefficient measures how much two distributions overlap, and is defined as $\rho(p_k, p_l) = \int \sqrt{p_k(x) p_l(x)} \mathrm{d}x$. The Bhattacharyya distance is one of the most classical distance function used in the literature [45] [46], and is defined as $-\log \rho(p_k, p_l)$. A weighted version of the distance is defined as [36].

$$Bhat(k,l) = -\min(\pi_k, \pi_l) \log \rho(p_k, p_l).$$

As we will see in Sections 4.1 and 4.2, *Bhat* performs very well in many situations, but can fail in the presence of a noise component, since it does not satisfy the Noise property. In **Figure 4(c)**, *Bhat* would have merged the blue and the red component.

**4) Kullback-Leibler information**

When one distribution is approximated by another, information is lost. The amount of lost information can be measured by the Kullback-Leibler information. For two probability measures $P_k$ and $P_l$, the Kullback-Leibler information, *I*, and Kullback-Leibler divergence, *J* are defined as [47]

$$I(k,l) = \int p_k(x) \log \frac{p_k(x)}{p_l(x)} \mathrm{d}x$$

$$J(k,l) = I(k,l) + I(l,k) = \int \left(p_k(x) - p_l(x)\right) \log \frac{p_k(x)}{p_l(x)} \mathrm{d}x$$

Since the Kullback-Leibler information is not symmetric, it is less suited for hybrid clustering. The Kullback-Leibler divergence was evaluated as dissimilarity measure for hybrid clustering by [21], but in the unweighted version. We therefore define a dissimilarity measure based on the Kullback-Leibler divergence as

$$KLdiv(k,l) = \min(\pi_k, \pi_l) \int \left(p_k(\boldsymbol{x}) - p_l(\boldsymbol{x})\right) \log \frac{p_k(\boldsymbol{x})}{p_l(\boldsymbol{x})} \mathrm{d}\boldsymbol{x}.$$

*A new dissimilarity measure*:

*KLdiv* does not satisfy the Noise property, and we therefore propose a symmetric dissimilarity measure based on the symmetrisation of the Kullback-Leibler information by the minimum function. The new dissimilarity measure is denoted by *KLinf*, and for two probability measures $P_k$ and $P_l$, it is defined as

$$KLinf(k,l) = \min(\pi_k, \pi_l) \min\left(\int p_k(\boldsymbol{x}) \log \frac{p_k(\boldsymbol{x})}{p_l(\boldsymbol{x})} \mathrm{d}\boldsymbol{x}, \int p_l(\boldsymbol{x}) \log \frac{p_l(\boldsymbol{x})}{p_k(\boldsymbol{x})} \mathrm{d}\boldsymbol{x}\right).$$

*KLinf* satisfies all our proposed properties in Section 3.2 and we therefore expect it to perform well in situations where these properties are desirable.

## 4. Illustration of the Dissimilarity Measure's Impact on the Final Clustering

We here illustrate how the choice of dissimilarity measure impacts the final clustering in view of the proposed properties. GMM is used to estimate the density, and BIC and AIC are used to estimate the number of components, corresponding to the number of subclusters. It does not suggest that this is the best partitional clustering technique, but for illustrational purposes it is appropriate, since it is well-known, relatively fast, and easily available. A pre-defined number of clusters is used as stopping rule since the focus of this article is the dissimilarity measure and not the stopping rule. The results for *wSE* are not displayed since its shortcomings are illustrated by *SE* and *JS*.

We compare the performances of the dissimilarity measures by the difference between the *misclassification rate*, which is dependent on the dissimilarity measure, and the *minimum misclassification rate*, which is defined by the partitional clustering (see Section 2.1). The misclassification rate is defined as the proportion of the observations wrongly clustered by the final clustering. To compare the dissimilarity measures, we define true cluster membership according to the samples' origin, as illustrated in Figure 5, which deviates from the usual concept of cluster. However, defining cluster membership by the relation between observations will favour some dissimilarity measures over others by construction, according to how relation is defined.

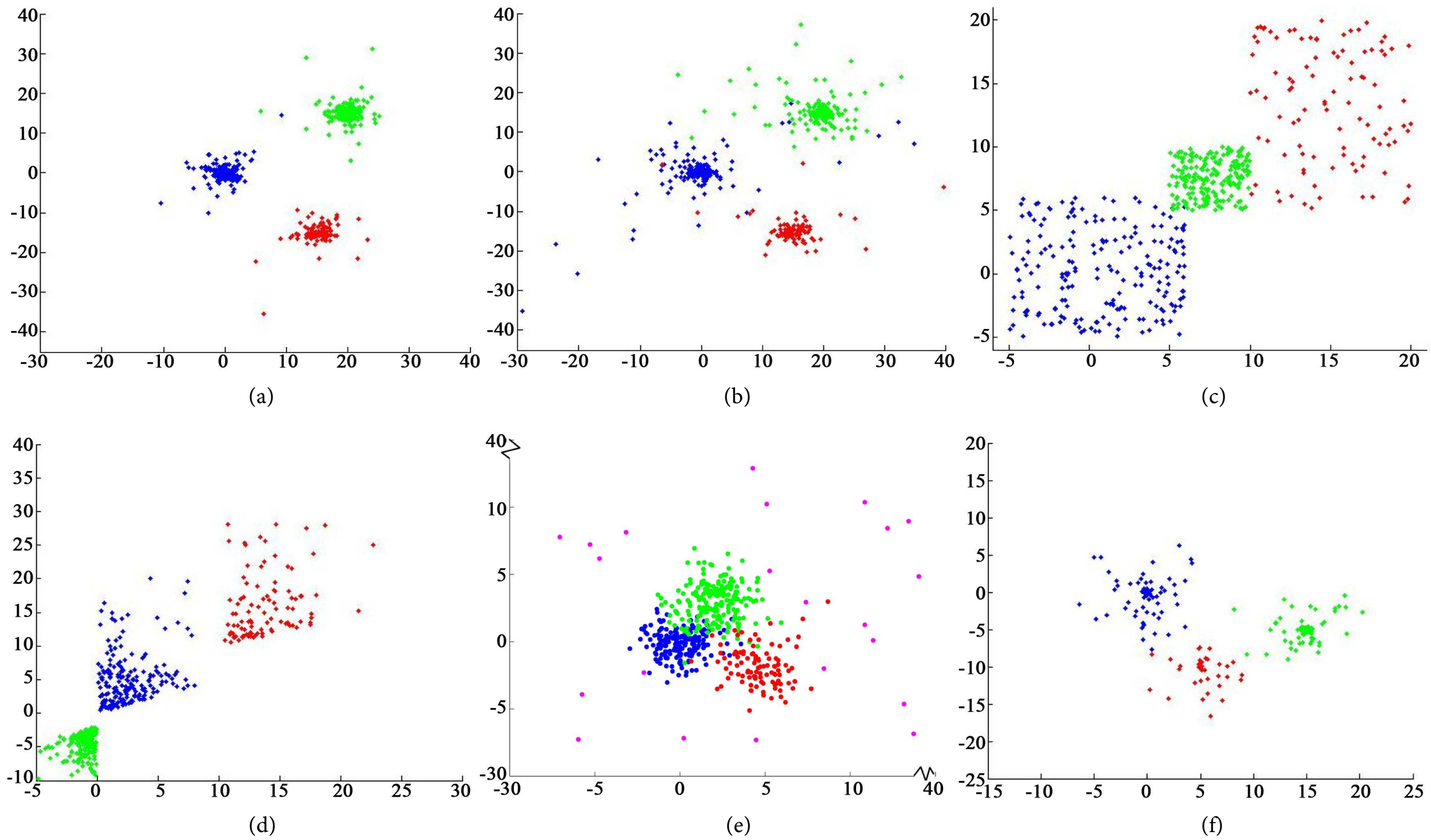

Figure 5. Examples of random samples of 500 observations drawn from different types of two-dimensional distributions. The colours indicate which cluster each sample is drawn from. (a) Student's *t*, $\nu = 2$; (b) Cauchy (Student's *t*, $\nu = 1$; (c) Uniform; (d) Gamma; (e) Gaussian with uniform noise; (f) Gaussian-Laplace mixture.

### 4.1. Simulation Study

We here present a simulation study for six types of distributions (for details, see Appendix C), shown in **Figure 5**. The wide variety in distributions and their deviance from the model for partitional clustering illustrate the relevance of the proposed properties. For each random sampling, a partitional clustering is performed by GMM, where the number of components is estimated by BIC and AIC. The hierarchical clustering is then performed according to each dissimilarity measure, until three clusters is reached. **Figure 6** shows the entire merging process. The mean values of 100 repetitions and their 95% confidence intervals are shown in **Figure 7**. There are eight varieties for each distribution: BIC (blue) or AIC (red) for the number of components, small (open) or large (filled) data sets, and dimension 2 (circles) or 3 (triangles). The noise observations in **Figure 5(e)** are not counted as misclassified, regardless of which cluster they are assigned to. As discussed in Section 2, both partitional clustering and the stopping criterion impact the final clustering, and it is possible that the results would have been different with another partitional clustering technique on the same data set.

From the above study we see that no dissimilarity measure performs uniformly best. We have observed some phenomena for different dissimilarity measures. The three dissimilarity measures that satisfy the Equality, Orthogonality and Outlier properties (*Bhat*, *KLdiv*, and *KLinf*) have overall lower misclassification than those that do not satisfy the aforementioned properties (*JS*, *SE*, and *Err*). When the observations are generated from a Student's $t$ distribution with 2 degrees of freedom, the clusters are well separated, symmetric and dense (see **Figure 5(a)**), but still, *JS*, *SE*, and *Err* fail to perform well. The poor performance of *SE* is shown in **Figure 6**. The misclassification rates for *JS*, *SE*, and *Err* are higher using AIC than using BIC because the number of components is higher with AIC, and consequently, the subcluster weights are smaller. For most distributions, there are no clear distinctions among *Bhat*, *KLdiv*, and *KLinf*, with some exceptions. For the Gaussian-Laplace mixture distribution, an example of a heavy-tailed distribution, *KLinf* performs better than *KLdiv*. This can be explained by investigating at which rate the dissimilarity measure changes according to changes in the constant $a$ described in the Mode property; $KLinf \to \infty$ faster than $KLdiv \to \infty$ as $a \to 0$ (see Appendix A for details). For the Gaussian with uniform noise distribution, *Bhat* and *KLdiv*, which do not satisfy the noise property, perform worse than *KLinf*. *Bhat* also does not perform well for the Cauchy distribution due to the noise-like observations from the heavy tails; The sparsely scattered observations with different origins result in *one* component with large scatter and low density. The AIC and the Kullback-Leibler information measure the fit of the model to the true distribution in a similar manner [28]. However, this will not favour the Kullback-Leibler based dissimilarity measures because they measure the fit between *components*.

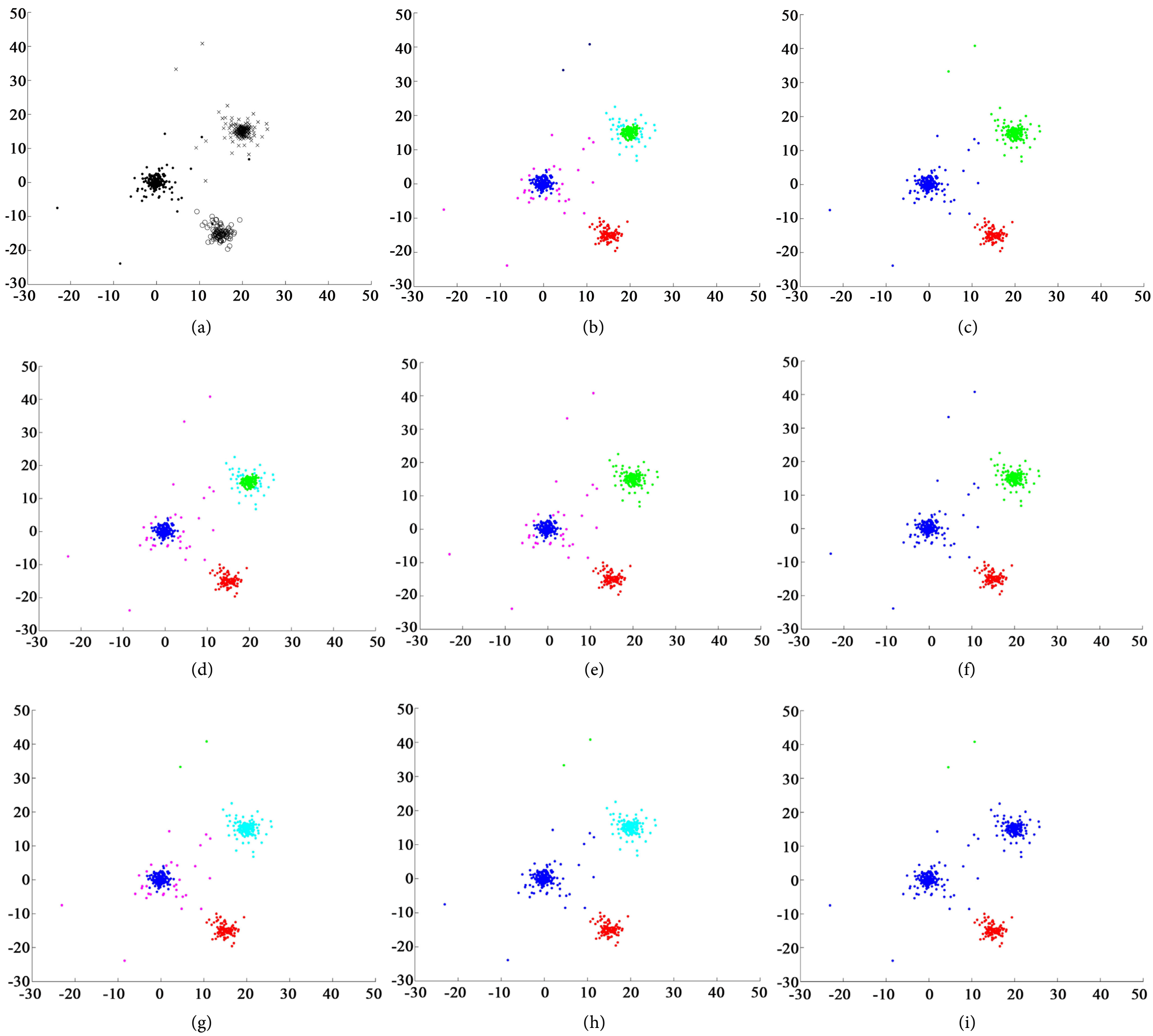

**Figure 6.** A random sample from Student's *t* distribution with 2 degrees of freedom. (a) True cluster membership (dots, circles, crosses); (b) MAP-component membership for a six-component GMM fitted to the observations; (c) The best merging in terms of minimum misclassification rate; (d)-(f) and (g)-(i) Hierarchical merging using *KLinf* and *SE*, respectively. (d) First merging step; *KLinf*; (e) Second step; *KLinf*; (f) Final merging; *KLinf*; (g) First merging step; *SE*; (h) Second step; *SE*; (i) Final merging; *SE*.

Several other studies have investigated hybrid clustering by simulated data. We illustrate the performance of the dissimilarity measures for the simulated data described in [22], where GMM/BIC was used for partitioning in hybrid clustering, in [32] and [34], where GMM/BIC was used as reference for ICL and Student's *t* mixture fitting, respectively. **Figure 8** shows the results for the data from [34] (blue), [32] (red), and [22] (cyan). The data are generated by random sampling from the distributions described in the respective papers, with 100 repetitions. The trends that appeared in

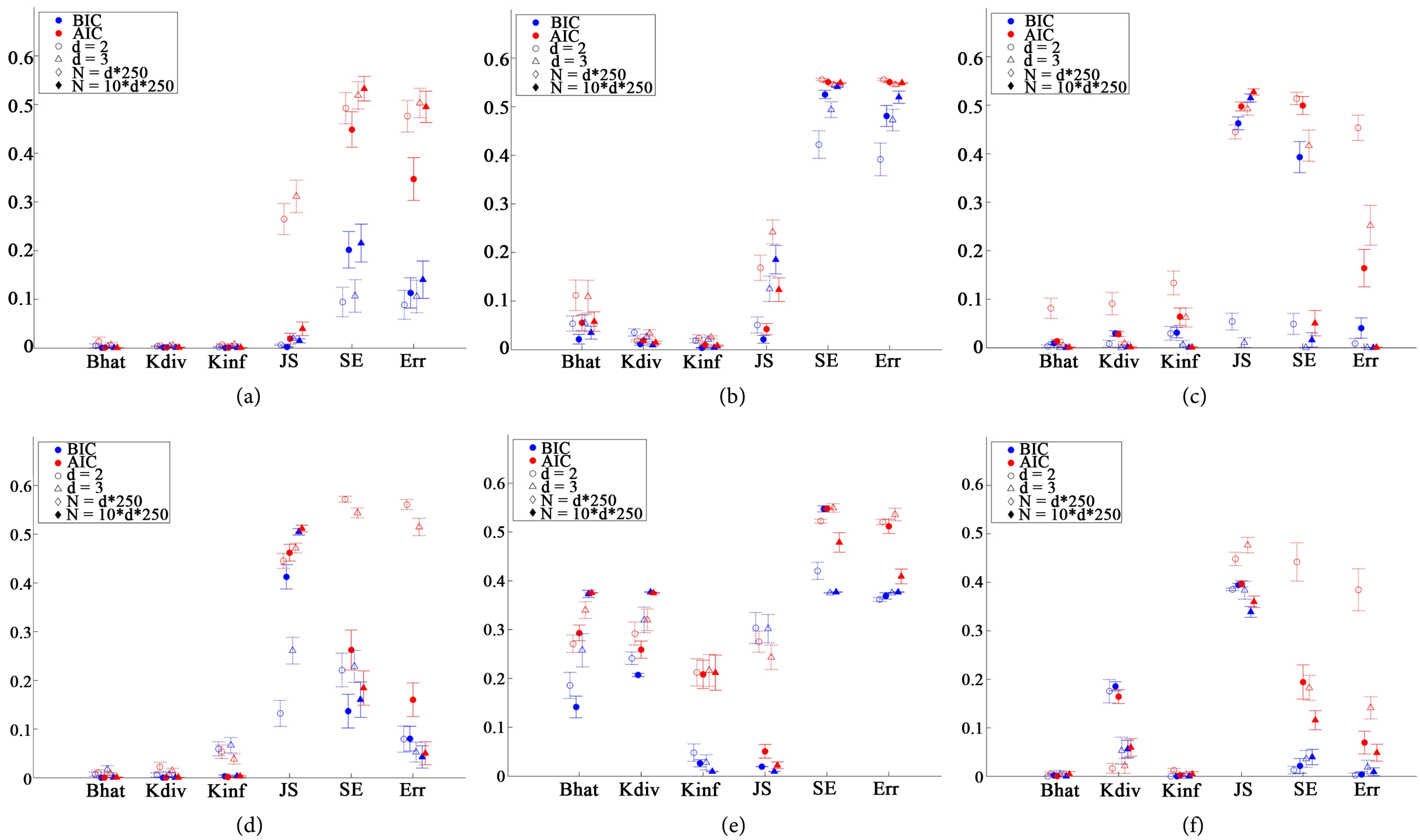

**Figure 7.** Difference between misclassification rate and minimum misclassification rate, and the 95% confidence intervals, for BIC and AIC, small and large data sets, dimension 2 and 3, for random samples from the distributions in **Figure 5**, obtained by GMM fitting and three final clusters. The number of components estimated by BIC and AIC varies for each sample. (a) Student's *t*, $\nu = 2$; (b) Cauchy (Student's *t*, $\nu = 1$; (c) Uniform; (d) Gamma; (e) Gaussian with uniform noise; (f) Gaussian-Laplace mixture.

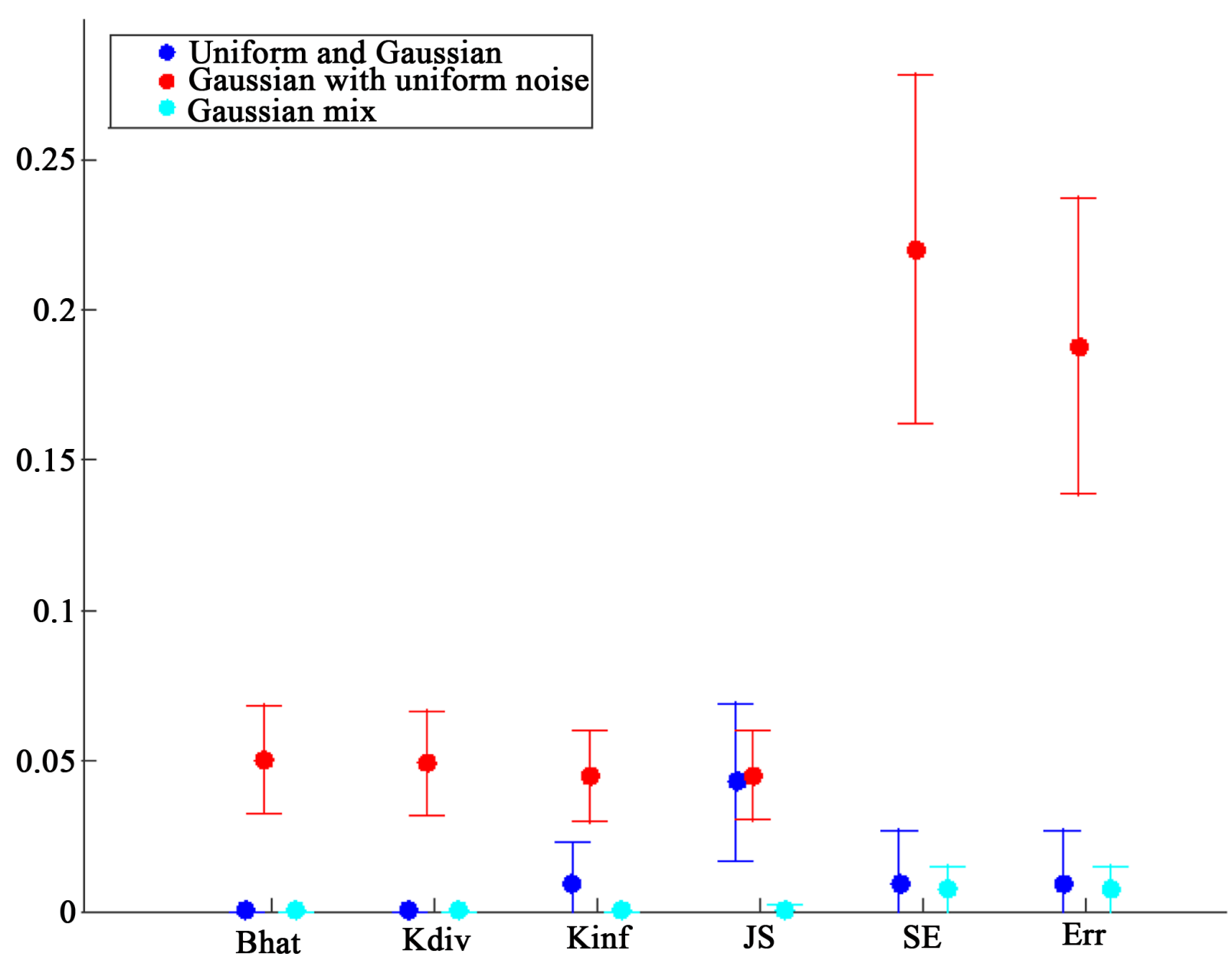

**Figure 8.** Difference between misclassification rate and minimum misclassification rate, and the 95% confidence intervals for simulated data, using GMM/BIC as partitional clustering.

Figure 7 are also seen in Figure 8 where no dissimilarity measure is uniformly best, but that *Bhat*, *KLdiv*, and *KLinf* perform better. For the Gaussian mixture distribution (cyan) described in [22], the choice of dissimilarity measure has very little impact on the final clustering. Interestingly, their proposed dissimilarity measure, *SE*, does not perform better than the others. *KLinf* fails more often than *Bhat* and *KLdiv* for the Uniform and Gaussian distribution.

## 4.2. Application to Real Data

**Wisconsin diagnostic breast cancer data:** The Wisconsin diagnostic breast cancer data, available from the UCI machine learning repository [30], consists of 569 observations in 30 dimensions and is labelled according to diagnosis, which defines the classes. Fraley and Raftery [48] used the three features selected by Mangasarian *et al.* [49] and achieved a misclassification rate of 5%. We have here used sequential forward feature selection. At each step, the feature that improves a quadratic discriminant classifier the most in terms of correctly classified observations is added. The process is stopped when adding a new feature results in an improvement of less than $10^{-6}$. The training and testing of the classifier is performed on separate sets, using cross-validation. The dimension of the resulting data sets varied from three to nine, and all features were selected at least once in 100 repetitions. Figure 9(a) shows an example with three dimensions chosen by the feature selection. The mean misclassification rate for *KLinf* and *Bhat* was 5%. The result is shown in Figure 10. As we have observed the phenomena in the simulated data, here also, *Bhat*, *KLdiv* and *KLinf* perform better than *JS*, *SE*, and *Err*. The Wisconsin breast cancer data is an example of data with higher dimensions, in this example, up to nine dimensions. The dissimilarity measures themselves are not limited by the number of dimensions and data sparsity, but indirectly they will suffer from bad density estimation.

**Yeast data:** A Student's *t* mixture distribution was used for partitional clustering of the Yeast data described in Section 1 (see Figure 3). The result for 10 repetitions is shown in Figure 10. We used Student's *t* mixture on the Yeast data to illustrate the use of another partitional clustering technique. The variation is due to slightly different estimated parameters for each repetition.

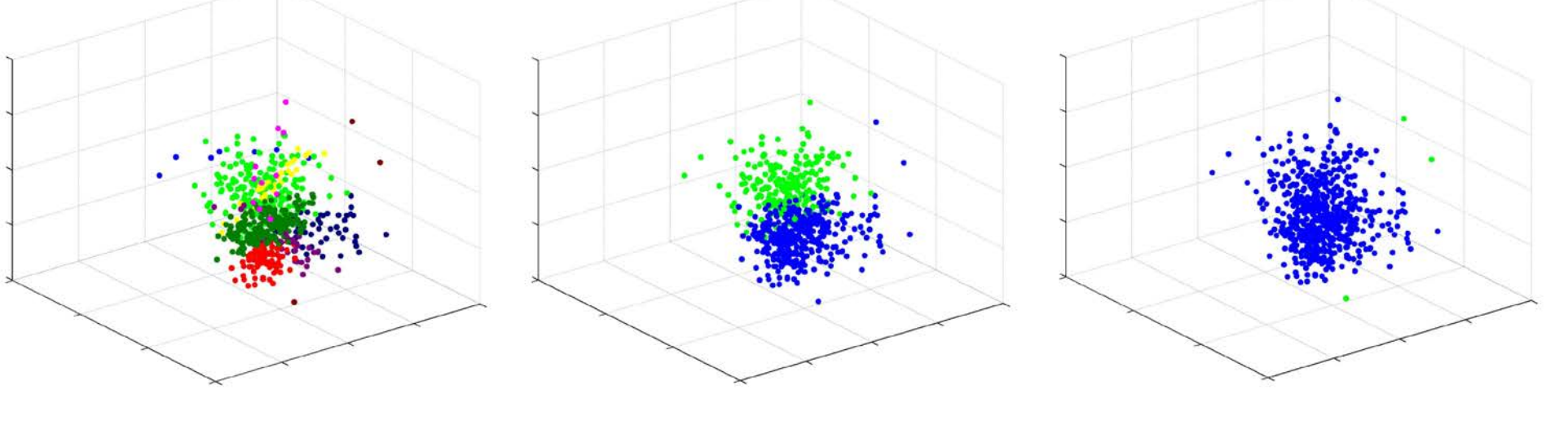

(a) (b) (c)

Figure 9. The Wisconsin diagnostic breast cancer data. (a) The initial partitioning, labelled according to MAP. Final clustering using (b) *Bhat* and (c) *SE*.

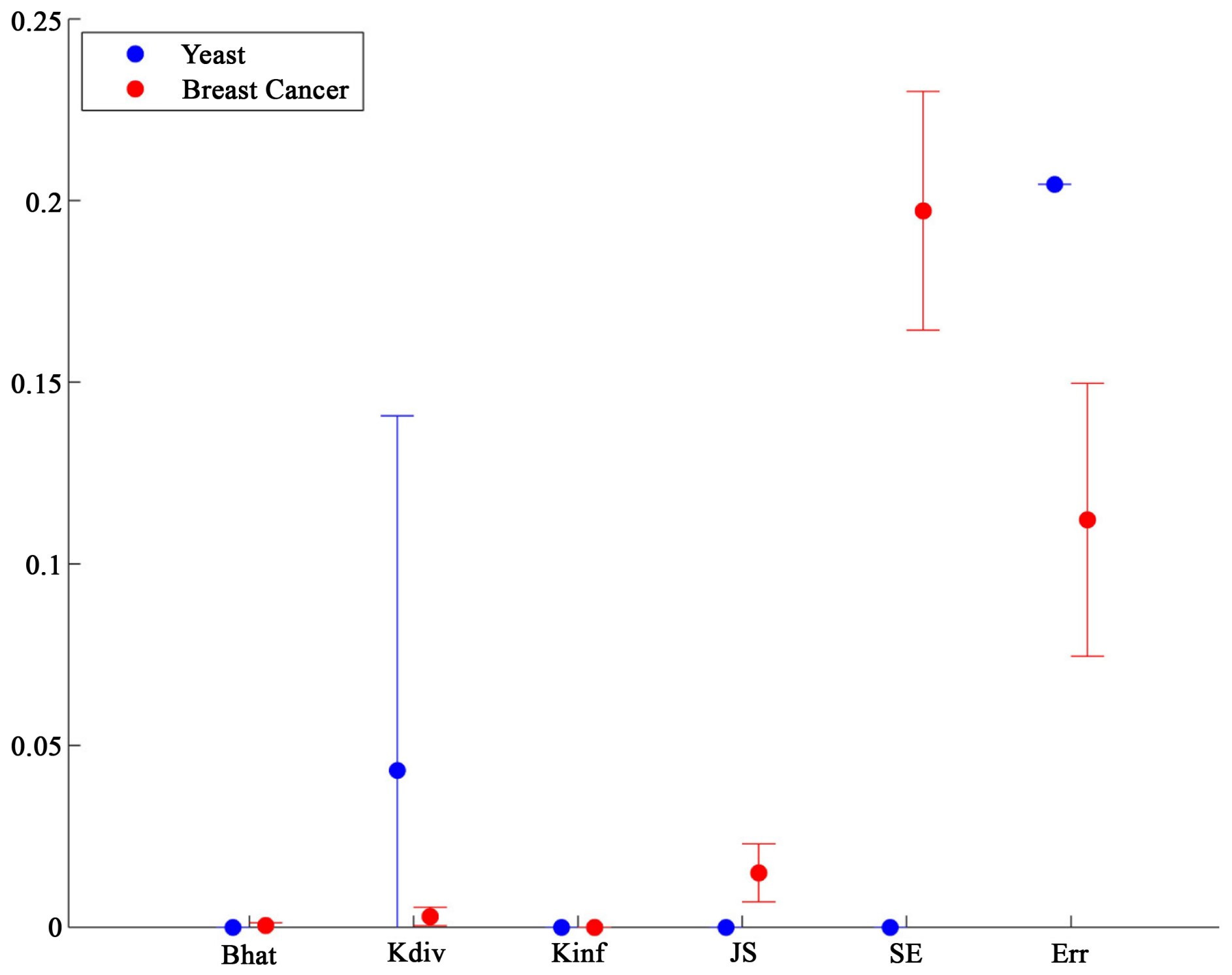

**Figure 10.** Difference between misclassification rate and minimum misclassification rate, and the 95% confidence intervals for Breast Cancer data using GMM as partitional clustering (red), and Yeast data using Student's *t* as partitional clustering (blue).

## 5. Discussion

In the literature, the choice of density-based dissimilarity measure for hybrid clustering has been given little attention, despite its heavy influence on the final clustering. We have proposed data-independent properties. The Equality and Orthogonality properties are presented as axiomatic for divergences [35], and although we will not claim that they are axiomatic for dissimilarity measures, it is indeed hard to imagine how they would be a drawback in hybrid clustering. The Outlier and the Noise properties ensure that the dissimilarity measure is robust to certain types of outlier and noise observations, meaning that whether outlier or noise observations are present or absent has little effect on the clustering. One can choose a partitional clustering that is less likely to produce small outlying subclusters (for example Student's *t* instead of GMM), but the possibility cannot be ruled out. Detection of outliers in pre-processing can reduce the necessity for robust dissimilarity measures, but requires extra knowledge about the data. The Mode property is valid only for the case of two Gaussian components with equally shaped covariance matrices, which is a special case. However, light-tailed models for heavy-tailed clusters are an important issue in hybrid clustering, since the need for hybrid clustering arises from a situation where the partitional clustering model does not fit the cluster distribution, and the Mode property can contribute in explaining a dissimilarity measure's behaviour under these circumstances.

The illustration of clustering results for different dissimilarity measures shows that the choice of dissimilarity measure can have significant impact on the final clustering. It also shows that the behaviour of a dissimilarity measure can be anticipated based on which properties it satisfies. However, the properties will not give a full description of a dissimilarity measure's behaviour, and analysis can give further insight. The different behaviour of *KLdiv* and *KLinf* for the Gaussian-Laplace mixture required further analysis, since both dissimilarity measures satisfied the Mode property. Data-independent properties can be used to choose the right dissimilarity measure, evaluate new dissimilarity measures, and modify existing ones. As an example of modification, the factor $\min(\pi_k, \pi_l)$ in the dissimilarity measures *Bhat*, *KLdiv*, and *KLinf* can be changed to, for example, $\log(\min(\pi_k, \pi_l))$. The dissimilarity measures will satisfy the Outlier property, but the behaviour with respect to how small the outlying component in Figure 4(b) must be before it is merged despite its near orthogonality, will change.

The final number of clusters is another legitimate issue in hybrid clustering. A property for dissimilarity measures when used for cluster number estimation, for instance by setting a threshold or searching for an "elbow" should be a topic for future studies. Figure 11 shows the values for *KLinf* and *SE* as the number of clusters goes from six to one for the sample in Figure 6. Both show an elbow at $C = 3$. The three-cluster solution is more appropriate for the *KLinf*-based merging than for the *SE*-based merging.

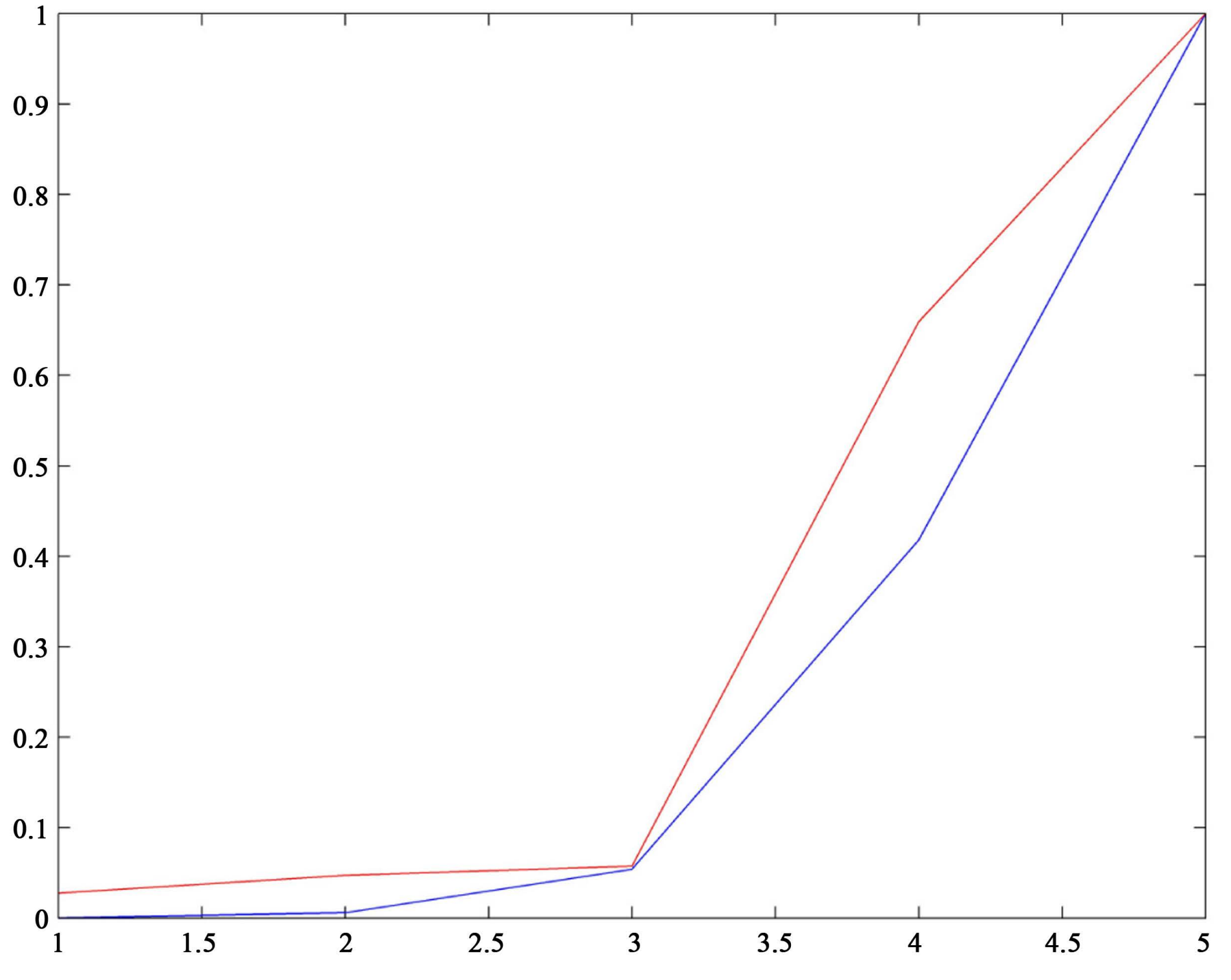

Figure 11. The normalised values of *KLinf* (red) and *SE* (blue). They both show an elbow at $C = 3$.

### 5.1. Main Contribution of the Article

We have proposed data-independent properties for density-based dissimilarity measures for hybrid clustering. Data-independent properties provide a tool for evaluating and comparing dissimilarity measures on a more general level than for particular data sets with known or user defined clusters.

The significance of the properties has been illustrated through studies on simulated and real data. We have evaluated several well-known dissimilarity measures in terms of the proposed properties. Besides, we have proposed a new dissimilarity measure, *KLinf*, that satisfies all the proposed properties. It shows good performance in both simulated and real data. However, we do not claim that *KLinf* is the *best* density-based dissimilarity measure for hybrid clustering; the concept of clustering is far too complex and diverse.

### 5.2. Concluding Remarks

Clustering is an exploratory tool for use in situations where the data have no ground truth. The diversity of clustering problems and the inherit subjectivity demands different dissimilarity measures for different problems. In this article, we have proposed specific properties, and those properties offer a tool to choose the most appropriate dissimilarity measure and to interpret the clustering result. In particular, a measure that does not fulfil the Equality or Orthogonality properties can give peculiar results, since it violates the most common concepts of "distance". Similarly, a dissimilarity measure that does not fulfil the Outlier property can lead to extreme results in many situations (see e.g. **Figure 9(c)**). The data presented in this article are in two or three dimensions, for illustration purposes, but for higher dimensions, the user cannot visually inspect the clustering result, and analytical knowledge is then crucial.

## References

[1] Hastie, T., Tibshirani, R. and Friedman, J. (2009) The Elements of Statistical Learning: Data mining, Inference, and Prediction. 2nd Edition, Springer Series in Statistics, Springer, New York.

[2] Jain, A.K., Murty, M.N. and Flynn, P.J. (1999) Data Clustering: A Review. *ACM Computing Surveys*, **31**, 264-323. http://dx.doi.org/10.1145/331499.331504

[3] Jain, A.K. (2010) Data Clustering: 50 Years beyond K-Means. *Pattern Recognition Letters*, **31**, 651-666. http://dx.doi.org/10.1016/j.patrec.2009.09.011

[4] Cox, D.R. (1957) Note on Grouping. *Journal of the American Statistical Association*, **52**, 543-547. http://dx.doi.org/10.1080/01621459.1957.10501411

[5] Fisher, W.D. (1958) On Grouping for Maximum Homogeneity. *Journal of the American Statistical Association*, **53**, 789-798. http://dx.doi.org/10.1080/01621459.1958.10501479

[6] Wong, M.A. (1982) A Hybrid Clustering Method for Identifying High-Density Clusters. *Journal of the American Statistical Association*, **77**, 841-847. http://dx.doi.org/10.1080/01621459.1982.10477896

[7] Goldberger, J. and Roweis, S.T. (2005) Hierarchical Clustering of a Mixture Model. Advances in Neural Information Processing Systems.

[8] Lin, C.-R. and Chen, M.-S. (2005) Combining Partitional and Hierarchical Algorithms for Robust and Efficient Data Clustering with Cohesion Self-Merging. *IEEE Transactions on Knowledge and Data Engineering*, **17**, 145-159. http://dx.doi.org/10.1109/TKDE.2005.21

[9] Liu, M., Jiang, X. and Kot, A.C. (2009) A Multi-Prototype Clustering Algorithm. *Pattern Recognition*, **42**, 689-698. http://dx.doi.org/10.1016/j.patcog.2008.09.015

[10] Murty, M.N. and Krishna, G. (1981) A Hybrid Clustering Procedure for Concentric and Chain-Like Clusters. *International Journal of Computer & Information Sciences*, **10**, 397-412. http://dx.doi.org/10.1007/BF00996137

[11] Patra, B.K., Nandi, S. and Viswanath, P. (2011) A Distance Based Clustering Method for Arbitrary Shaped Clusters in Large Datasets. *Pattern Recognition*, **44**, 2862-2870. http://dx.doi.org/10.1016/j.patcog.2011.04.027

[12] Vijaya, P.A., Murty, N.M. and Subramanian, D.K. (2006) Efficient Bottom-Up Hybrid Hierarchical Clustering Techniques for Protein Sequence Classification. *Pattern Recognition*, **39**, 2344-2355. http://dx.doi.org/10.1016/j.patcog.2005.12.001

[13] Viswanath, P. and Suresh Babu, V. (2009) Rough-DBSCAN: A Fast Hybrid Density Based Clustering Method for Large Data Sets. *Pattern Recognition Letters*, **30**, 1477-1488. http://dx.doi.org/10.1016/j.patrec.2009.08.008

[14] Zhang, T., Ramakrishnan, R. and Livny, M. (1996) BIRCH: An Efficient Data Clustering Method for Very Large Databases. *Proceedings of the* 1996 *ACM SIGMOD International Conference on Management of Data*, New York, 103-114. http://dx.doi.org/10.1145/233269.233324

[15] Zhong, S. and Ghosh, J. (2003) A Unified Framework for Model-Based Clustering. *Journal of Machine Learning Research*, **4**, 1001-1037.

[16] García-Escudero, L.A., Gordaliza, A., Matrán, C. and Mayo-Iscar, A. (2011) Exploring the Number of Groups in Robust Model-Based Clustering. *Statistics and Computing*, **21**, 585-599. http://dx.doi.org/10.1007/s11222-010-9194-z

[17] Arbelaitz, O., Gurrutxaga, I., Muguerza, J., Pérez, J.M. and Perona, I. (2013) An Extensive Comparative Study of Cluster Validity Indices. *Pattern Recognition*, **46**, 243-256. http://dx.doi.org/10.1016/j.patcog.2012.07.021

[18] Dubes, R. and Jain, A.K. (1976) Clustering Techniques: The User's Dilemma. *Pattern Recognition*, **8**, 247-260. http://dx.doi.org/10.1016/0031-3203(76)90045-5

[19] Kleinberg, J. (2003) An Impossibility Theorem for Clustering. *Advances in Neural Information Processing Systems* (*NIPS* 2002), **15**, 463-470.

[20] Puzicha, J., Hofmann, T. and Buhmann, J.M. (2000) A Theory of Proximity Based Clustering: Structure Detection by Optimization. *Pattern Recognition*, **33**, 617-634. http://dx.doi.org/10.1016/S0031-3203(99)00076-X

[21] Puzicha, J., Buhmann, J.M., Rubner, Y. and Tomasi, C. (1999) Empirical Evaluation of Dissimilarity Measures for Color and Texture. *The Proceedings of the Seventh IEEE International Conference on Computer Vision*, **2**, 1165-1172. http://dx.doi.org/10.1109/ICCV.1999.790412

[22] Baudry, J.-P., Raftery, A.E., Celeux, G., Lo, K. and Gottardo, R. (2010) Combining Mixture Components for Clustering. *Journal of Computational and Graphical Statistics*, **9**, 332-353. http://dx.doi.org/10.1198/jcgs.2010.08111

[23] Karypis, G., Han, E.-H. and Kumar, V. (1999) Chameleon: Hierarchical Clustering Using Dynamic Modeling. *Computer*, **32**, 68-75. http://dx.doi.org/10.1109/2.781637

[24] Huber, M.F., Bailey, T., Durrant-Whyte, H. and Hanebeck, U.D. (2008) On Entropy Approximation for Gaussian Mixture Random Vectors. *IEEE International Conference on*

*Multisensor Fusion and Integration for Intelligent Systems*, August 2008, 181-188. http://dx.doi.org/10.1109/MFI.2008.4648062

[25] Maz'ya, V. and Schmidt, G. (1996) On Approximate Approximations Using Gaussian Kernels. *IMA Journal of Numerical Analysis*, 13-29. http://dx.doi.org/10.1093/imanum/16.1.13

[26] Dempster, A.P., Laird, N.M. and Rubin, D.B. (1977) Maximum Likelihood from Incomplete Data via the EM Algorithm. *Journal of the Royal Statistical Society*, *Series B* (*Methodological*), **39**, 1-38.

[27] Schwarz, G. (1978) Estimating the Dimension of a Model. *The Annals of Statistics*, **6**, 461-464. http://dx.doi.org/10.1214/aos/1176344136

[28] Akaike, H. (1974) A New Look at the Statistical Model Identification. *IEEE Transactions on Automatic Control*, **19**, 716-723. http://dx.doi.org/10.1109/TAC.1974.1100705

[29] McLachlan, G. and Peel, D. (2000) Finite Mixture Models. Wiley Series in Probability and Statistics, John Wiley & Sons, Inc. http://dx.doi.org/10.1002/0471721182

[30] Bache, K. and Lichman, M. (2013) UCI Machine Learning Repository.

[31] Franczak, B.C., Browne, R.P. and McNicholas, P.D. (2014) Mixtures of Shifted Asymmetric Laplace Distributions. *IEEE Transactions on Pattern Analysis and Machine Intelligence*, **36**, 1149-1157. http://dx.doi.org/10.1109/TPAMI.2013.216

[32] Peel, D. and McLachlan, G.J. (2000) Robust Mixture Modelling Using the t Distribution. *Statistics and Computing*, **10**, 339-348. http://dx.doi.org/10.1023/A:1008981510081

[33] R. Tibshirani, G. Walther, and T. Hastie, \Estimating the number of clusters in a data set via the gap statistic," Journal of the Royal Statistical Society: Series B (Statistical Methodology), vol. 63, pp. 411{423, Jan. 2001.

[34] Biernacki, C., Celeux, G. and Govaert, G. (2000) Assessing a Mixture Model for Clustering with the Integrated Completed Likelihood. *IEEE Transactions on Pattern Analysis and Machine Intelligence*, **22**, 719-725. http://dx.doi.org/10.1109/34.865189

[35] Ali, S.M. and Silvey, S.D. (1966) A General Class of Coefficients of Divergence of One Distribution from Another. *Journal of the Royal Statistical Society*, *Series B* (*Methodological*), **28**, 131-142.

[36] Calderero, F. and Marques, F. (2008) General Region Merging Approaches Based on Information Theory Statistical Measures. 15*th IEEE International Conference on Image Processing*, 3016-3019.

[37] Basseville, M. (1989) Distance Measures for Signal Processing and Pattern Recognition. *Signal Processing*, **18**, 349-369. http://dx.doi.org/10.1016/0165-1684(89)90079-0

[38] Csiszár, I. (1967) Information-Type Measures of Difference of Probability Distributions and Indirect Observations. *Studia Scientiarum Mathematicarum Hungarica*, **2**, 299-318.

[39] Pardo, L. (2006) Statistical Inference Based on Divergence Measures. Vol. 185 of Statistics, Chapman and Hall/CRC.

[40] Berger, A. (1953) On Orthogonal Probability Measures. *Proceedings of the American Mathematical Society*, **4**, 800-806. http://dx.doi.org/10.1090/S0002-9939-1953-0056868-5

[41] Farcomeni, A. (2013) Robust Constrained Clustering in Presence of Entry-Wise Outliers. *Technometrics*, **56**, 102-111. http://dx.doi.org/10.1080/00401706.2013.826148

[42] Huber, P.J. and Ronchetti, E.M. (2009) Robust Statistics. Wiley Series in Probability and Statistics. 2nd Edition, John Wiley & Sons, Inc., Hoboken.

[43] Shannon, C.E. (1948) A Mathematical Theory of Communication, Part I. *The Bell System Technical Journal*, **27**, 379-423. http://dx.doi.org/10.1002/j.1538-7305.1948.tb01338.x

[44] Lin, J. (1991) Divergence Measures Based on the Shannon Entropy. *IEEE Transactions on Information Theory*, **37**, 145-151. http://dx.doi.org/10.1109/18.61115

[45] Bhattacharyya, A. (1943) On a Measure of Divergence between Two Statistical Populations Defined by Their Probability Distributions. *Bulletin of the Calcutta Mathematical Society*, **35**, 99-109.

[46] Kailath, T. (1967) The Divergence and Bhattacharyya Distance Measures in Signal Selection. *IEEE Transactions on Communication Technology*, **15**, 52-60. http://dx.doi.org/10.1109/TCOM.1967.1089532

[47] Kullback, S. and Leibler, R.A. (1951) On Information and Sufficiency. *The Annals of Mathematical Statistics*, **22**, 79-86. http://dx.doi.org/10.1214/aoms/1177729694

[48] Fraley, C. and Raftery, A.E. (2002) Model-Based Clustering, Discriminant Analysis, and Density Estimation. *Journal of the American Statistical Association*, **97**, 611-631. http://dx.doi.org/10.1198/016214502760047131

[49] Mangasarian, O.L., Street, W.N. and Wolberg, W.H. (1995) Breast Cancer Diagnosis and Prognosis via Linear Programming. *Operations Research*, **43**, 570-577. http://dx.doi.org/10.1287/opre.43.4.570

[50] Furman, E. (2008) On a Multivariate Gamma Distribution. *Statistics & Probability Letters*, **78**, 2353-2360. http://dx.doi.org/10.1016/j.spl.2008.02.012

# Appendix

## Appendix A

**Some preliminaries:**

1) It is straightforward to see that the dissimilarity measures in Section 3.3 are symmetric.

2) The integral is taken over $\mathcal{X}$ unless otherwise noted. Here $\mathcal{X}$ denotes the support of the distribution function associated with the random variable *X*. Let $p_k(\boldsymbol{x})$ be the pdf of subcluster *k*, and $\pi_k$ denote the weight, $\pi_i > 0$. Let $w_k = \pi_k/(\pi_k + \pi_l)$ and $w_l = \pi_l/(\pi_k + \pi_l)$. $D(k,l)$ denotes the dissimilarity measure between the two subclusters *k* and *l*.

3) Orthogonality is defined as [40]: $p_k(\boldsymbol{x})$ and $p_l(\boldsymbol{x})$ are orthogonal if for any $\epsilon > 0$, there exist $S_k$ and $S_l$ such that

$$\int_{S_k} p_k(\boldsymbol{x})\mathrm{d}\boldsymbol{x} > 1-\epsilon, \int_{S_k} p_l(\boldsymbol{x})\mathrm{d}\boldsymbol{x} < \epsilon, \int_{S_l} p_l(\boldsymbol{x})\mathrm{d}\boldsymbol{x} > 1-\epsilon \text{ and } \int_{S_l} p_k(\boldsymbol{x})\mathrm{d}\boldsymbol{x} < \epsilon,$$

where $S_k \cup S_l = \mathcal{X}$.

4) Note that $\int_{S_k} p_l(\boldsymbol{x})\mathrm{d}\boldsymbol{x} < \epsilon$ for any $\epsilon > 0$ implies that for any $\delta > 0$, $p_l(\boldsymbol{x}) < \delta$ almost everywhere (a.e.) on $S_k$, since $p_l(\boldsymbol{x}) \geq 0$, and similarly $\int_{S_l} p_k(\boldsymbol{x})\mathrm{d}\boldsymbol{x} < \varepsilon$ for any $\epsilon > 0$ implies that for any $\delta > 0$, $p_k(\boldsymbol{x}) < \delta$ a.e. on $S_l$, since $p_k(\boldsymbol{x}) \geq 0$.

5) The logarithmic function is monotonic increasing, and consequently, $\log(a+b) > \log(a)$, if $b > 0$. Since $p_k(\boldsymbol{x})$ is a pdf, $\int p_k(\boldsymbol{x}) = 1$.

6) If a dissimilarity measure does not satisfy either the Equality or the Orthogonality property, the Mode property will not be satisfied, since $a \to 1$ gives $p_k(\boldsymbol{x}) = p_l(\boldsymbol{x})$, and $a \to 0$ gives $p_k(\boldsymbol{x}) \perp p_l(\boldsymbol{x})$. If a dissimilarity measure is an *f*-divergence, the Mode property is automatically satisfied. This is not necessarily true for modified *f*-divergences.

Next, we investigate whether the dissimilarity measures in Section 3.3 satisfy the properties proposed in Section 3.2.

**Shannon entropy based measures**

1) *Unweighted version*, *SE*: Recall that

$$\begin{aligned} SE(k,l) = &-\int \pi_k p_k(\boldsymbol{x}) \log\left(\pi_k p_k(\boldsymbol{x}) + \pi_l p_l(\boldsymbol{x})\right)\mathrm{d}\boldsymbol{x} \\ &-\int \pi_l p_l(\boldsymbol{x}) \log\left(\pi_k p_k(\boldsymbol{x}) + \pi_l p_l(\boldsymbol{x})\right)\mathrm{d}\boldsymbol{x} \\ &+\int \pi_k p_k(\boldsymbol{x}) \log \pi_k p_k(\boldsymbol{x})\mathrm{d}\boldsymbol{x} + \int \pi_l p_l(\boldsymbol{x}) \log \pi_l p_l(\boldsymbol{x})\mathrm{d}\boldsymbol{x}, \end{aligned}$$

and note that $\max(SE) = 0$ and $\min(SE) = -\infty$. We have the following results:

**Equality:** Set $\pi_l = \pi_k$ and $p_l(\boldsymbol{x}) = p_k(\boldsymbol{x})$. We now have

$$\begin{aligned} SE(k,l) &= -\int \pi_k p_k(\boldsymbol{x}) \log\left(\pi_k p_k(\boldsymbol{x}) + \pi_k p_k(\boldsymbol{x})\right)\mathrm{d}\boldsymbol{x} \\ &\quad -\int \pi_k p_k(\boldsymbol{x}) \log\left(\pi_k p_k(\boldsymbol{x}) + \pi_k p_k(\boldsymbol{x})\right)\mathrm{d}\boldsymbol{x} \\ &\quad +\int \pi_k p_k(\boldsymbol{x}) \log \pi_k p_k(\boldsymbol{x})\mathrm{d}\boldsymbol{x} + \int \pi_k p_k(\boldsymbol{x}) \log \pi_k p_k(\boldsymbol{x})\mathrm{d}\boldsymbol{x} \\ &= -2\int \pi_k p_k(\boldsymbol{x}) \log 2\mathrm{d}\boldsymbol{x} - 2\int \pi_k p_k(\boldsymbol{x}) \log \pi_k p_k(\boldsymbol{x})\mathrm{d}\boldsymbol{x} + \int 2\pi_k p_k(\boldsymbol{x}) \log \pi_k p_k(\boldsymbol{x})\mathrm{d}\boldsymbol{x} \\ &= -2\pi_k \log 2 \int p_k \mathrm{d}\boldsymbol{x} = -2\pi_k \log 2. \end{aligned}$$

This implies that $SE$ does not satisfy the Equality property, since $\min(SE) = -\infty$.

**Orthogonality:** Let $p_k(\mathbf{x}) \perp p_l(\mathbf{x})$, and $S_k \cap S_l = \varnothing$. Then we can write

$$\begin{aligned} SE(k,l) = &-\int_{S_k} \pi_k p_k(\mathbf{x}) \log\left(\pi_k p_k(\mathbf{x}) + \pi_l p_l(\mathbf{x})\right) \mathrm{d}\mathbf{x} + \int_{S_k} \pi_k p_k(\mathbf{x}) \log \pi_k p_k(\mathbf{x}) \mathrm{d}\mathbf{x} \\ &- \int_{S_k} \pi_l p_l(\mathbf{x}) \log\left(\pi_k p_k(\mathbf{x}) + \pi_l p_l(\mathbf{x})\right) \mathrm{d}\mathbf{x} + \int_{S_k} \pi_l p_l(\mathbf{x}) \log \pi_l p_l(\mathbf{x}) \mathrm{d}\mathbf{x} \\ &- \int_{S_l} \pi_k p_k(\mathbf{x}) \log\left(\pi_k p_k(\mathbf{x}) + \pi_l p_l(\mathbf{x})\right) \mathrm{d}\mathbf{x} + \int_{S_l} \pi_l p_l(\mathbf{x}) \log \pi_l p_l(\mathbf{x}) \mathrm{d}\mathbf{x} \\ &- \int_{S_l} \pi_l p_l(\mathbf{x}) \log\left(\pi_k p_k(\mathbf{x}) + \pi_l p_l(\mathbf{x})\right) \mathrm{d}\mathbf{x} + \int_{S_l} \pi_k p_k(\mathbf{x}) \log \pi_k p_k(\mathbf{x}) \mathrm{d}\mathbf{x}. \end{aligned}$$

Let

$$\left| -\int_{S_k} \pi_k p_k(\mathbf{x}) \log\left(\pi_k p_k(\mathbf{x}) + \pi_l p_l(\mathbf{x})\right) \mathrm{d}\mathbf{x} + \int_{S_k} \pi_k p_k(\mathbf{x}) \log \pi_k p_k(\mathbf{x}) \mathrm{d}\mathbf{x} \right| < \epsilon_1,$$

$$\left| -\int_{S_k} \pi_l p_l(\mathbf{x}) \log\left(\pi_k p_k(\mathbf{x}) + \pi_l p_l(\mathbf{x})\right) \mathrm{d}\mathbf{x} \right| < \epsilon_2,$$

$$\left| \int_{S_k} \pi_l p_l(\mathbf{x}) \log\left(\pi_l p_l(\mathbf{x})\right) \mathrm{d}\mathbf{x} \right| < \epsilon_3,$$

$$\left| -\int_{S_l} \pi_l p_l(\mathbf{x}) \log\left(\pi_k p_k(\mathbf{x}) + \pi_l p_l(\mathbf{x})\right) \mathrm{d}\mathbf{x} + \int_{S_l} \pi_l p_l(\mathbf{x}) \log \pi_l p_l(\mathbf{x}) \mathrm{d}\mathbf{x} \right| < \epsilon_4,$$

$$\left| -\int_{S_l} \pi_l p_l(\mathbf{x}) \log\left(\pi_k p_k(\mathbf{x}) + \pi_l p_l(\mathbf{x})\right) \mathrm{d}\mathbf{x} \right| < \epsilon_5,$$

$$\left| \int_{S_l} \pi_l p_l(\mathbf{x}) \log\left(\pi_k p_k(\mathbf{x})\right) \mathrm{d}\mathbf{x} \right| < \epsilon_6.$$

Since $p_l(\mathbf{x}) < \delta$ a.e. on $S_k$ and $p_k(\mathbf{x}) < \delta$ a.e. on $S_l$ for any $\delta > 0$, there exists a $\delta$ such that $\sum_{i=1}^{6} \epsilon_i < \epsilon$. Consequently, $|SE(k,l)| < \epsilon$. Since this is true for any $\epsilon > 0$, and $\max(SE) = 0$, it follows that $SE$ satisfies the Orthogonality property.

**Outlier:** Without loss of generality, let $\min(\pi_k, \pi_l) = \pi_k < \delta$. Let

$$\left| -\int \pi_l p_l(\mathbf{x}) \log\left(\pi_k p_k(\mathbf{x}) + \pi_l p_l(\mathbf{x})\right) \mathrm{d}\mathbf{x} + \int \pi_l p_l(\mathbf{x}) \log \pi_l p_l(\mathbf{x}) \mathrm{d}\mathbf{x} \right| < \epsilon_1,$$

$$\left| -\int \pi_k p_k(\mathbf{x}) \log\left(\pi_k p_k(\mathbf{x}) + \pi_l p_l(\mathbf{x})\right) \mathrm{d}\mathbf{x} \right| < \epsilon_2,$$

$$\left| \int \pi_k p_k(\mathbf{x}) \log \pi_k p_k(\mathbf{x}) \mathrm{d}\mathbf{x} \right| < \epsilon_3.$$

For any $\epsilon > 0$, there exists a $\delta > 0$ such that $\sum_{i=1}^{3} \epsilon_i < \epsilon$. Consequently, $|SE(k,l)| < \epsilon$, and since $\min(SE) = -\infty$, it follows that $SE$ does not satisfy the Outlier property.

**Noise:** Let $\max\left(p_k(\mathbf{x})\right) < \delta$, and let

$$\left| -\int \pi_l p_l(\mathbf{x}) \log\left(\pi_k p_k(\mathbf{x}) + \pi_l p_l(\mathbf{x})\right) \mathrm{d}\mathbf{x} + \int \pi_l p_l(\mathbf{x}) \log \pi_l p_l(\mathbf{x}) \mathrm{d}\mathbf{x} \right| < \epsilon_1,$$

$$\left| -\int \pi_k p_k(\mathbf{x}) \log\left(\pi_k p_k(\mathbf{x}) + \pi_l p_l(\mathbf{x})\right) \mathrm{d}\mathbf{x} \right| < \epsilon_2,$$

$$\left| \int \pi_k p_k(\mathbf{x}) \log \pi_k p_k(\mathbf{x}) \mathrm{d}\mathbf{x} \right| < \epsilon_3.$$

For any $\epsilon > 0$, there exists a $\delta > 0$ such that $\sum_{i=1}^{3} \epsilon_i < \epsilon$. Consequently, $|SE(k,l)| < \epsilon$, and since $\min(SE) = -\infty$, it follows that $SE$ does not satisfy the Noise

property.

**Mode:** Since *SE* does not satisfy the Equality property, it will not satisfy the Mode property. □

2) *Weighted version*, *wSE*: Recall that

$$wSE(k,l) = -\pi_k \int \left(p_k(\boldsymbol{x}) + p_l(\boldsymbol{x})\right) \log\left(p_k(\boldsymbol{x}) + p_l(\boldsymbol{x})\right) \mathrm{d}\boldsymbol{x}$$
$$- \pi_l \int \left(p_k(\boldsymbol{x}) + p_l(\boldsymbol{x})\right) \log\left(p_k(\boldsymbol{x}) + p_l(\boldsymbol{x})\right) \mathrm{d}\boldsymbol{x}$$
$$+ \pi_k \int p_k(\boldsymbol{x}) \log p_k \mathrm{d}\boldsymbol{x} + \pi_l \int p_l(\boldsymbol{x}) \log p_l(\boldsymbol{x}) \mathrm{d}\boldsymbol{x},$$

and note that $\min(wSE) = -\infty$ and $\max(wSE) = 0$. We have the following results:

**Equality:** Let $\pi_l = \pi_k$ and $p_l(\boldsymbol{x}) = p_k(\boldsymbol{x})$. We then have

$$wSE(k,k) = -(\pi_k + \pi_k) \int \left(p_k(\boldsymbol{x}) + p_k(\boldsymbol{x})\right) \log\left(p_k(\boldsymbol{x}) + p_k(\boldsymbol{x})\right) \mathrm{d}\boldsymbol{x}$$
$$+ \pi_k \int p_k(\boldsymbol{x}) \log p_k(\boldsymbol{x}) \mathrm{d}\boldsymbol{x} + \pi_k \int p_k(\boldsymbol{x}) \log p_k(\boldsymbol{x}) \mathrm{d}\boldsymbol{x}$$
$$= -2\pi_k \int 2 p_k(\boldsymbol{x}) \log 2 p_k(\boldsymbol{x}) \mathrm{d}\boldsymbol{x} + 2\pi_k \int p_k(\boldsymbol{x}) \log p_k(\boldsymbol{x}) \mathrm{d}\boldsymbol{x}$$
$$= -4\pi_k \int p_k(\boldsymbol{x}) \left(\log 2 + \log p_k(\boldsymbol{x})\right) \mathrm{d}\boldsymbol{x} + 2\pi_k \int p_k(\boldsymbol{x}) \log p_k(\boldsymbol{x}) \mathrm{d}\boldsymbol{x}$$
$$= -2\pi_k \int p_k(\boldsymbol{x}) \log p_k(\boldsymbol{x}) \mathrm{d}\boldsymbol{x} - 4\pi_k \log 2 \int p_k(\boldsymbol{x}) \mathrm{d}\boldsymbol{x}$$
$$= -2\pi_k \int p_k(\boldsymbol{x}) \log p_k(\boldsymbol{x}) \mathrm{d}\boldsymbol{x} - 4\pi_k \log 2.$$

Since $wSE(k,k)$ attains its minimum only if $p_k(\boldsymbol{x})$ has infinite entropy, the Equality property is not satisfied.

**Orthogonality:** Let $p_k(\boldsymbol{x}) \perp p_l(\boldsymbol{x})$, and $S_k \cap S_l = \varnothing$. Then we can write

$$wSE(k,l) = -\pi_k \int_{S_k} \left(p_k(\boldsymbol{x}) + p_l(\boldsymbol{x})\right) \log\left(p_k(\boldsymbol{x}) + p_l(\boldsymbol{x})\right) \mathrm{d}\boldsymbol{x} + \pi_k \int_{S_k} p_k(\boldsymbol{x}) \log p_k(\boldsymbol{x}) \mathrm{d}\boldsymbol{x}$$
$$- \pi_l \int_{S_k} \left(p_k(\boldsymbol{x}) + p_l(\boldsymbol{x})\right) \log\left(p_k(\boldsymbol{x}) + p_l(\boldsymbol{x})\right) \mathrm{d}\boldsymbol{x} + \pi_l \int_{S_k} p_l(\boldsymbol{x}) \log p_l(\boldsymbol{x}) \mathrm{d}\boldsymbol{x}$$
$$- \pi_k \int_{S_l} \left(p_k(\boldsymbol{x}) + p_l(\boldsymbol{x})\right) \log\left(p_k(\boldsymbol{x}) + p_l(\boldsymbol{x})\right) \mathrm{d}\boldsymbol{x} + \pi_k \int_{S_l} p_k(\boldsymbol{x}) \log p_k(\boldsymbol{x}) \mathrm{d}\boldsymbol{x}$$
$$- \pi_l \int_{S_l} \left(p_k(\boldsymbol{x}) + p_l(\boldsymbol{x})\right) \log\left(p_k(\boldsymbol{x}) + p_l(\boldsymbol{x})\right) \mathrm{d}\boldsymbol{x} + \pi_l \int_{S_l} p_l(\boldsymbol{x}) \log p_l(\boldsymbol{x}) \mathrm{d}\boldsymbol{x}.$$

Let

$$-\pi_k \int_{S_k} \left(p_k(\boldsymbol{x}) + p_l(\boldsymbol{x})\right) \log\left(p_k(\boldsymbol{x}) + p_l(\boldsymbol{x})\right) \mathrm{d}\boldsymbol{x} + \pi_k \int_{S_k} p_k(\boldsymbol{x}) \log p_k(\boldsymbol{x}) \mathrm{d}\boldsymbol{x} = \epsilon_1,$$

$$-\pi_l \int_{S_k} p_l(\boldsymbol{x}) \log\left(p_k(\boldsymbol{x}) + p_l(\boldsymbol{x})\right) \mathrm{d}\boldsymbol{x} = \epsilon_2,$$

$$+\pi_l \int_{S_k} p_l(\boldsymbol{x}) \log p_l(\boldsymbol{x}) \mathrm{d}\boldsymbol{x} = \epsilon_3,$$

$$-\pi_l \int_{S_l} \left(p_k(\boldsymbol{x}) + p_l(\boldsymbol{x})\right) \log\left(p_k(\boldsymbol{x}) + p_l(\boldsymbol{x})\right) \mathrm{d}\boldsymbol{x} + \pi_l \int_{S_l} p_l(\boldsymbol{x}) \log p_l(\boldsymbol{x}) \mathrm{d}\boldsymbol{x} = \epsilon_4,$$

$$-\pi_k \int_{S_l} p_k(\boldsymbol{x}) \log\left(p_k(\boldsymbol{x}) + p_l(\boldsymbol{x})\right) \mathrm{d}\boldsymbol{x} = \epsilon_5,$$

$$+\pi_k \int_{S_l} p_k(\boldsymbol{x}) \log p_k(\boldsymbol{x}) \mathrm{d}\boldsymbol{x} = \epsilon_6.$$

Consequently,

$$wSE(k,l) = \sum_{i=1}^{6} \epsilon_i - \pi_l \int_{S_k} p_k(\boldsymbol{x}) \log\left(p_k(\boldsymbol{x}) + p_l(\boldsymbol{x})\right) \mathrm{d}\boldsymbol{x}$$
$$- \pi_k \int_{S_l} p_l(\boldsymbol{x}) \log\left(p_k(\boldsymbol{x}) + p_l(\boldsymbol{x})\right) \mathrm{d}\boldsymbol{x}$$

and attains its maximum only if the two latter terms are equal to zero. It follows that $wSE$ does not satisfy the Orthogonality property.

**Outlier:** Without loss of generality, let $\min(\pi_k, \pi_l) = \pi_k < \delta$. Let

$$-\pi_k \int (p_k(\boldsymbol{x}) + p_l(\boldsymbol{x})) \log (p_k(\boldsymbol{x}) + p_l(\boldsymbol{x})) \mathrm{d}\boldsymbol{x} = \epsilon_1,$$

$$\pi_k \int p_k(\boldsymbol{x}) \log p_k \mathrm{d}\boldsymbol{x} = \epsilon_2.$$

Consequently,

$$\begin{aligned} wSE(k,l) = \epsilon_1 + \epsilon_2 - \pi_l \int (p_k(\boldsymbol{x}) + p_l(\boldsymbol{x})) \log (p_k(\boldsymbol{x}) + p_l(\boldsymbol{x})) \mathrm{d}\boldsymbol{x} \\ + \pi_l \int p_l(\boldsymbol{x}) \log p_l(\boldsymbol{x}) \mathrm{d}\boldsymbol{x} \end{aligned}$$

and attains its minimum only if the two latter terms are infinite. It follows that $wSE$ does not satisfy the Outlier property.

**Noise:** Let $\max(p_k(\boldsymbol{x})) < \delta$, and let

$$-\pi_k \int p_k(\boldsymbol{x}) \log (p_k(\boldsymbol{x}) + p_l(\boldsymbol{x})) \mathrm{d}\boldsymbol{x} = \epsilon_1,$$

$$-\pi_l \int (p_k(\boldsymbol{x}) + p_l(\boldsymbol{x})) \log (p_k(\boldsymbol{x}) + p_l(\boldsymbol{x})) \mathrm{d}\boldsymbol{x} + \pi_l \int p_l(\boldsymbol{x}) \log p_l(\boldsymbol{x}) \mathrm{d}\boldsymbol{x} = \epsilon_2,$$

$$\pi_k \int p_k(\boldsymbol{x}) \log p_k(\boldsymbol{x}) \mathrm{d}\boldsymbol{x} = \epsilon_3.$$

Consequently, $wSE(k,l) = \sum_{i=1}^{3} \epsilon_i - \pi_k \int p_l(\boldsymbol{x}) \log p_l(\boldsymbol{x}) \mathrm{d}\boldsymbol{x}$, and attains its minimum only if the entropy of $p_l(\boldsymbol{x})$ is infinite. It follows that $wSE$ does not satisfy the Noise property.

**Mode:** Since $wSE$ does not satisfy the Equality and the Orthogonality properties, it will not satisfy the Mode property. □

3) *Jensen-Shannon dissimilarity measure*, *JS*: Recall that

$$\begin{aligned} JS(k,l) = -\int (w_k p_k(\boldsymbol{x}) + w_l p_l(\boldsymbol{x})) \log (w_k p_k(\boldsymbol{x}) + w_l p_l(\boldsymbol{x})) \mathrm{d}\boldsymbol{x} \\ + w_k \int p_k(\boldsymbol{x}) \log p_k(\boldsymbol{x}) \mathrm{d}\boldsymbol{x} + w_l \int p_l(\boldsymbol{x}) \log p_l(\boldsymbol{x}) \mathrm{d}\boldsymbol{x}, \end{aligned}$$

and note that $\min(JS) = 0$ and $\max(JS) = \log 2$.

**Equality:** Set $\pi_l = \pi_k$ and $p_l(\boldsymbol{x}) = p_k(\boldsymbol{x})$, and hence, $w_k = w_l = 1/2$. We then have

$$\begin{aligned} JS(k,k) &= -\int (w_k p_k(\boldsymbol{x}) + w_k p_k(\boldsymbol{x})) \log (w_k p_k(\boldsymbol{x}) + w_k p_k(\boldsymbol{x})) \mathrm{d}\boldsymbol{x} \\ &\quad + w_k \int p_k(\boldsymbol{x}) \log p_k(\boldsymbol{x}) \mathrm{d}\boldsymbol{x} + w_k \int p_k(\boldsymbol{x}) \log p_k(\boldsymbol{x}) \mathrm{d}\boldsymbol{x} \\ &= -2w_k \int p_k(\boldsymbol{x}) \log 2w_k p_k(\boldsymbol{x}) \mathrm{d}\boldsymbol{x} + 2w_k \int p_k(\boldsymbol{x}) \log p_k(\boldsymbol{x}) \mathrm{d}\boldsymbol{x} \\ &= -\int p_k(\boldsymbol{x}) \log p_k(\boldsymbol{x}) \mathrm{d}\boldsymbol{x} + \int p_k(\boldsymbol{x}) \log p_k(\boldsymbol{x}) \mathrm{d}\boldsymbol{x} = 0. \end{aligned}$$

Since $\min(JS) = 0$, the Equality property is satisfied.

**Orthogonality:** Let $p_k(\boldsymbol{x}) \perp p_l(\boldsymbol{x})$, and $S_k \cap S_l = \emptyset$. Then we can write

$$\begin{aligned} JS(k,l) = -\int_{S_k} (w_k p_k(\boldsymbol{x}) + w_l p_l(\boldsymbol{x})) \log (w_k p_k(\boldsymbol{x}) + w_l p_l(\boldsymbol{x})) \mathrm{d}\boldsymbol{x} \\ + w_k \int_{S_k} p_k(\boldsymbol{x}) \log p_k(\boldsymbol{x}) \mathrm{d}\boldsymbol{x} + w_l \int_{S_k} p_l(\boldsymbol{x}) \log p_l(\boldsymbol{x}) \mathrm{d}\boldsymbol{x} \\ - \int_{S_l} (w_k p_k(\boldsymbol{x}) + w_l p_l(\boldsymbol{x})) \log (w_k p_k(\boldsymbol{x}) + w_l p_l(\boldsymbol{x})) \mathrm{d}\boldsymbol{x} \\ + w_k \int_{S_l} p_k(\boldsymbol{x}) \log p_k(\boldsymbol{x}) \mathrm{d}\boldsymbol{x} + w_l \int_{S_l} p_l(\boldsymbol{x}) \log p_l(\boldsymbol{x}) \mathrm{d}\boldsymbol{x}, \end{aligned}$$

and

$$-\int_{S_k} w_k p_k(\boldsymbol{x}) \log\left(w_k p_k(\boldsymbol{x}) + w_l p_l(\boldsymbol{x})\right) \mathrm{d}\boldsymbol{x} + w_k \int_{S_k} p_k(\boldsymbol{x}) \log p_k(\boldsymbol{x}) \mathrm{d}\boldsymbol{x}$$
$$= -w_k \log w_k \int_{S_k} p_k(\boldsymbol{x}) \mathrm{d}\boldsymbol{x} + \epsilon_1 = -w_k \log w_k \left(1 - \epsilon_2^*\right) + \epsilon_1 = -w_k \log w_k + \epsilon_1 + \epsilon_2$$

$$-\int_{S_k} w_l p_l(\boldsymbol{x}) \log\left(w_k p_k(\boldsymbol{x}) + w_l p_l(\boldsymbol{x})\right) \mathrm{d}\boldsymbol{x} = \epsilon_3$$

$$w_l \int_{S_k} p_l(\boldsymbol{x}) \log p_l(\boldsymbol{x}) \mathrm{d}\boldsymbol{x} = \epsilon_4$$

$$-\int_{S_l} w_l p_l(\boldsymbol{x}) \log\left(w_l p_l(\boldsymbol{x}) + w_k p_k(\boldsymbol{x})\right) \mathrm{d}\boldsymbol{x} + w_l \int_{S_l} p_l(\boldsymbol{x}) \log p_l(\boldsymbol{x}) \mathrm{d}\boldsymbol{x}$$
$$= -w_l \log w_l \int_{S_l} p_l(\boldsymbol{x}) \mathrm{d}\boldsymbol{x} + \epsilon_5 = -w_l \log w_l \left(1 - \epsilon_6^*\right) + \epsilon_5 = -w_l \log w_l + \epsilon_5 + \epsilon_6$$

$$-\int_{S_l} w_k p_k(\boldsymbol{x}) \log\left(w_l p_l(\boldsymbol{x}) + w_k p_k(\boldsymbol{x})\right) \mathrm{d}\boldsymbol{x} = \epsilon_7$$

$$w_k \int_{S_l} p_k(\boldsymbol{x}) \log p_k(\boldsymbol{x}) \mathrm{d}\boldsymbol{x} = \epsilon_8$$

Consequently, $JS(k,l) = \sum_{i=1}^{8} \epsilon_i - w_k \log w_k - w_l \log w_l$. Since $JS(k,l)$ will attain its maximum value only if $w_k = w_l$, the Orthogonality property is not satisfied.

**Outlier:** Let $\pi_l = \pi_k$. Then $w_k$ and $w_l$ are constant regardless of the value of $\pi_k$, and

$$JS(k,l) = -\int \left(w_k p_k(\boldsymbol{x}) + w_l p_l(\boldsymbol{x})\right) \log\left(w_k p_k(\boldsymbol{x}) + w_l p_l(\boldsymbol{x})\right) \mathrm{d}\boldsymbol{x}$$
$$+ w_k \int p_k(\boldsymbol{x}) \log p_k(\boldsymbol{x}) \mathrm{d}\boldsymbol{x} + w_l \int p_l(\boldsymbol{x}) \log p_l(\boldsymbol{x}) \mathrm{d}\boldsymbol{x}$$

does not satisfy the Outlier property.

**Noise:** Let $\max\left(p_k(\boldsymbol{x})\right) < \delta$, and let

$$-\int w_l p_l(\boldsymbol{x}) \log\left(w_k p_k(\boldsymbol{x}) + w_l p_l(\boldsymbol{x})\right) \mathrm{d}\boldsymbol{x} + w_l \int p_l(\boldsymbol{x}) \log p_l(\boldsymbol{x}) \mathrm{d}\boldsymbol{x} = -w_l \log w_l + \epsilon_1,$$

$$-\int w_k p_k(\boldsymbol{x}) \log\left(w_k p_k(\boldsymbol{x}) + w_l p_l(\boldsymbol{x})\right) \mathrm{d}\boldsymbol{x} = \epsilon_2,$$

$$w_k \int p_k(\boldsymbol{x}) \log p_k(\boldsymbol{x}) \mathrm{d}\boldsymbol{x} = \epsilon_3.$$

Consequently, $JS(k,l) = \sum_{i=1}^{3} \epsilon_i - w_l \log w_l$. Since the latter term attains zero only if $w_l = 1$ or $w_l = 0$, the Noise property is not satisfied.

**Mode:** Since *JS* does not satisfy the Orthogonality property, it will not satisfy the Mode property. □

**Probability of misclassification**

1) *Probability of misclassification*, *Err*: Recall that

$$Err(k,l) = 1 - \int \min\left(w_k p_k(\boldsymbol{x}), w_l p_l(\boldsymbol{x})\right) \mathrm{d}\boldsymbol{x},$$

and note that $\min(Err) = 1/2$ and $\max(Err) = 1$. Note also that *Err* is an *f*-divergence, and hence satisfies the **Equality**, **Orthogonality** and **Mode** properties.

**Outlier:** Without loss of generality, let $\min(\pi_k, \pi_l) = \pi_k < \delta$. For any $\epsilon > 0$ there exists a $\delta > 0$ such that

$$1 - \int \min\left(w_k p_k(\boldsymbol{x}), w_l p_l(\boldsymbol{x})\right) \mathrm{d}\boldsymbol{x} > 1 - \epsilon,$$

and hence $Err(k,l)$ does not satisfy the Outlier property, since $\min(Err) = 1/2$.

**Noise:** Let $\max\left(p_k\left(\boldsymbol{x}\right)\right) < \delta$. For any $\epsilon > 0$ there exists a $\delta > 0$ such that

$$1 - \int \min\left(w_k p_k\left(\boldsymbol{x}\right), w_l p_l\left(\boldsymbol{x}\right)\right) \mathrm{d}\boldsymbol{x} > 1 - \epsilon,$$

and hence $Err\left(k,l\right)$ does not satisfy the Noise property, since $\min\left(Err\right) = 1/2$. □

**Bhattacharyya distance**

1) *Weighted Bhattacharyya distance*, *Bhat*: Recall that

$$Bhat\left(k,l\right) = -\min\left(\pi_k, \pi_l\right) \log \int \sqrt{p_k\left(\boldsymbol{x}\right) p_l\left(\boldsymbol{x}\right)} \mathrm{d}\boldsymbol{x},$$

and note that $\min\left(Bhat\right) = 0$ and $\max\left(Bhat\right) = \infty$. Note also that the Bhattacharyya distance, $-\log \int \sqrt{p_k\left(\boldsymbol{x}\right) p_l\left(\boldsymbol{x}\right)} \mathrm{d}\boldsymbol{x}$, is an *f*-divergence, and hence it is straightforward to show that *Bhat* satisfies the **Equality** and **Orthogonality** properties.

**Outlier:** Without loss of generality, let $\min\left(\pi_k, \pi_l\right) = \pi_k < \delta$. Then, for any $\epsilon > 0$, there exists a $\delta > 0$ such that

$$Bhat\left(k,l\right) = -\min\left(\pi_k, \pi_l\right) \log \int \sqrt{p_k\left(\boldsymbol{x}\right) p_l\left(\boldsymbol{x}\right)} \mathrm{d}\boldsymbol{x} < \epsilon$$

for $-\log \int \sqrt{p_k\left(\boldsymbol{x}\right) p_l\left(\boldsymbol{x}\right)} \mathrm{d}\boldsymbol{x} < \infty$, and hence, $Bhat\left(k,l\right)$ satisfies the Outlier property.

**Noise:** Let $\max\left(p_k\left(\boldsymbol{x}\right)\right) < \delta$. Then, for any $\epsilon > 0$, there exists a $\delta > 0$ such that

$$Bhat\left(k,l\right) = -\min\left(\pi_k, \pi_l\right) \log \int \sqrt{p_k\left(\boldsymbol{x}\right) p_l\left(\boldsymbol{x}\right)} \mathrm{d}\boldsymbol{x} > \epsilon,$$

and hence, $Bhat\left(k,l\right)$ does not satisfy the Noise property.

**Mode:** Since the Bhattacharyya distance is an *f*-divergence, it is straightforward to show that *Bhat* satisfies the Mode property. However, it is of interest to know how *Bhat* behaves as a function of *a*. The Bhattacharrya distance between two normal distributions is

$$\frac{1}{4}\left(\boldsymbol{\mu}_l - \boldsymbol{\mu}_k\right)^{\mathrm{T}} \left(\Sigma_k + \Sigma_l\right)^{-1} \left(\boldsymbol{\mu}_l - \boldsymbol{\mu}_k\right) + \frac{1}{2} \log \frac{\left|\left(\Sigma_k + \Sigma_l\right)/2\right|}{\sqrt{\left|\Sigma_k \Sigma_l\right|}},$$

and hence in the special case of $\boldsymbol{\mu}_k = \boldsymbol{\mu}_l$ and $\Sigma_k = a\Sigma_l$, we have

$$Bhat\left(k,l\right) = \min\left(\pi_k, \pi_l\right) \frac{1}{2} \log \frac{\left|\left(\Sigma + a\Sigma\right)/2\right|}{\sqrt{\left|a\Sigma\Sigma\right|}} = \frac{1}{2} \log \frac{\left(\frac{a+1}{2}\right)^d \left|\Sigma\right|}{a^{d/2} \left|\Sigma\right|} = \frac{1}{2} \log \frac{\left(a+1\right)^d}{2^d a^{d/2}}.$$

□

**Kullback-Leibler information**

*Kullback-Leibler divergence and Kullback-Leibler information*: Recall that

$$I\left(k,l\right) = \int p_k\left(\boldsymbol{x}\right) \log p_k\left(\boldsymbol{x}\right) \mathrm{d}\boldsymbol{x} - \int p_k\left(\boldsymbol{x}\right) \log p_l\left(\boldsymbol{x}\right) \mathrm{d}\boldsymbol{x}$$

$$J\left(k,l\right) = I\left(k,l\right) + I\left(l,k\right)$$

$$KLinf\left(k,l\right) = \min\left(\pi_k, \pi_l\right) \min\left(I\left(k,l\right), I\left(l,k\right)\right)$$

$$KLdiv\left(k,l\right) = \min\left(\pi_k, \pi_l\right) J\left(k,l\right)$$

and note that $\min\left(I\right) = 0$ and $\max\left(I\right) = \infty$, and hence $\min\left(KLinf\right) = 0$, $\max\left(KLinf\right) = \infty$, $\min\left(KLdiv\right) = 0$, $\max\left(KLdiv\right) = \infty$. Note also that $I$ and $J$ are *f*- divergences, and hence it is straightforward to show that *KLinf* and *KLdiv* satisfy the **Equality** and **Orthogonality** properties.

**Outlier:** Without loss of generality, let $\min(\pi_k, \pi_l) = \pi_k < \delta$. Then, for any $\epsilon > 0$, there exists a $\delta > 0$ such that

$$KLdiv(k,l) = -\min(\pi_k, \pi_l) J(k,l) < \epsilon$$

$$KLinf(k,l) = -\min(\pi_k, \pi_l) I(k,l) < \epsilon$$

for $I(k,l) < \infty$ and $J(k,l) < \infty$, and hence, *KLdiv* and *KLinf* satisfy the Outlier property.

**Noise:** Let $\max(p_k(\boldsymbol{x})) < \delta$. Then, for any $\epsilon > 0$, there exists a $\delta > 0$ such that

$$I(k,l) = \int p_k(\boldsymbol{x}) \log p_k(\boldsymbol{x}) \mathrm{d}\boldsymbol{x} - \int p_k(\boldsymbol{x}) \log p_l(\boldsymbol{x}) \mathrm{d}\boldsymbol{x} < \epsilon$$

$$I(l,k) = \int p_l(\boldsymbol{x}) \log p_l(\boldsymbol{x}) \mathrm{d}\boldsymbol{x} - \int p_l(\boldsymbol{x}) \log p_k(\boldsymbol{x}) \mathrm{d}\boldsymbol{x} > \epsilon$$

and hence, $KLdiv(k,l)$ does not satisfy the Noise property. However, $KLinf(k,l)$ satisfies the Noise property.

**Mode:** Since both $I(k,l)$ and $J(k,l)$ are *f*-divergences, it is straightforward to show that *KLdiv* and *KLinf* satisfy the Mode property. However, it is of interest to know how the dissimilarity measures behave as a function of *a*. The Kullback-Leibler information between two normal distributions is

$$\frac{1}{2}\left( tr\left(\Sigma_l^{-1}\Sigma_k\right) + \left(\boldsymbol{\mu}_l - \boldsymbol{\mu}_k\right)^{\mathrm{T}} \Sigma_l^{-1}\left(\boldsymbol{\mu}_l - \boldsymbol{\mu}_k\right) - d - \log\frac{|\Sigma_l|}{|\Sigma_k|} \right),$$

and hence in the special case of $\boldsymbol{\mu}_k = \boldsymbol{\mu}_l$ and $\Sigma_k = a\Sigma_l$, we have

$$\frac{1}{2}\left( tr\left(\Sigma^{-1}\Sigma/a\right) - d - \log\frac{|\Sigma|}{a|\Sigma|} \right),$$

which result in

$$I(k,l) = \frac{1}{2}\left(\frac{d}{a} - d - \log\frac{1}{a}\right) \text{ and } I(l,k) = \frac{1}{2}\left(ad - d - \log a\right).$$

□

## Appendix B

The probability density functions displayed in Figure 4 are Gaussian mixtures,

$$\phi(x) = \sum_{i=1}^{K} \pi_i \phi_i(x)$$

where $\phi_i(x)$ is a Gaussian distribution with mean $\mu_i$ and variance $\sigma_i$, and $\sum_{i=1}^{K} \pi_i = 1$. In the hybrid clustering context we have the following scenario: The coloured lines represent the current subclusters and the next step is to merge the two subclusters with lowest dissimilarity. The figures illustrate the relevance of the proposed properties.

(a) The distribution consists of three components with equal variance, but different component weights. We have that $|\mu_1 - \mu_2| < |\mu_2 - \mu_3|$, and therefore $d(\phi_1, \phi_2) < d(\phi_2, \phi_3)$ for any *f*-divergence independent of component weights, e.g. the (original) Bhattach-

aryya distance and the (original) Kullback-Leibler information. This is due to the fourth *f*-property, and also applies for other dissimilarity measures that fulfil this property. From left to right:

Blue: $\mu_1 = -3, \sigma_1 = 1, \pi_1 = 0.475$

Red: $\mu_2 = 0, \sigma_2 = 1, \pi_2 = 0.475$

Green: $\mu_3 = 3.1, \sigma_3 = 1, \pi_3 = 0.05$

(b) The distribution consists of three components, where one of the components is small and near orthogonal to the others. Orthogonality can be achieved by increasing the distance between the means, by reducing the variance, or by decreasing the size of the smallest component. In this example, $SE(\phi_1, \phi_2) < SE(\phi_2, \phi_3)$ and $Err(\phi_1, \phi_2) < Err(\phi_2, \phi_3)$, which is consistent with their violation of the Outlier property (see Section 3.2 and **Table 1**). This is true for any dissimilarity measure that fulfils the Orthogonality property but violates the Outlier property, as the size of the smallest component approaches zero. From left to right:

Blue: $\mu_1 = -1, \sigma_1 = 1, \pi_1 = 0.505$

Red: $\mu_2 = 4, \sigma_2 = 1, \pi_2 = 0.490$

Green: $\mu_3 = 10, \sigma_3 = 0.5, \pi_3 = 0.005$

(c) The distribution consists of four components, where the background noise consists of two components. In this example, $Bhat(\phi_1, \phi_2) < Bhat(\phi_1, \phi_3 + \phi_4)$, which is consistent with the violation of the Noise property (see Section 3.2 and **Table 1**). This is true for any dissimilarity measure that violates the Noise property, as the maximum of the noise density approaches zero, in this example, e.g., by increasing the variances of the two noise components.

Blue: $\mu_1 = -1.5, \sigma_1 = 1, \pi_1 = 0.332$

Red: $\mu_2 = 1.5, \sigma_2 = 1, \pi_2 = 0.332$

Green: $\pi_3 \phi(x \mid \mu_3, \sigma_3) + \pi_4 \phi(x \mid \mu_4, \sigma_4), \mu_3 = -15, \sigma_3 = 15, \pi_3 = 0.168, \mu_4 = 15, \sigma_4 = 15, \pi_4 = 0.168$

(d) The distribution consists of three components, where two of them have equal mean but unequal variance. In this example, $KLdiv(\phi_1, \phi_2) < KLdiv(\phi_2, \phi_3)$ but $KLinf(\phi_1, \phi_2) > KLinf(\phi_2, \phi_3)$. Both fulfil the Mode property, but they behave different as a function of *a*, where $\Sigma_k = a\Sigma_l$ (see Appendix A).

Blue: $\mu_1 = 0, \sigma_1 = 1, \pi_1 = 1/3$

Red: $\mu_2 = 3, \sigma_2 = 1, \pi_2 = 1/3$

Green: $\mu_3 = 3, \sigma_3 = 0.2, \pi_3 = 1/3$

(e) The distribution consists of four components, where the green and the pink component are identical, and therefore only the pink line is visible. In this example, $SE(\phi_1, \phi_2) < SE(\phi_3, \phi_4)$, which is consistent with *SE* violating the Equality property (see Section 3.2 and **Table 1**).

Blue: $\mu_1 = -2.9, \sigma_1 = 1, \pi_1 = 0.4$

Red: $\mu_2 = 0, \sigma_2 = 1, \pi_2 = 0.4$

Green: $\mu_3 = 2.5, \sigma_3 = 0.24, \pi_3 = 0.1$

Pink: $\mu_4 = 2.5, \sigma_4 = 0.24, \pi_4 = 0.1$

(f) This distribution consists of four components, where three of the components have the same weights, and are small relative to the first component. In this example $JS(\phi_1,\phi_2) < JS(\phi_2,\phi_3)$, and illustrates that a dissimilarity measure should be a function of the absolute values of the component weights, not only the pairwise relative weights.

Blue: $\mu_1 = -4, \sigma_1 = 0.75, \pi_1 = 2/3$

Red: $\mu_2 = 4, \sigma_2 = 0.75, \pi_2 = 1/9$

Green: $\mu_3 = 6, \sigma_3 = 0.75, \pi_3 = 1/9$

Pink: $\mu_4 = 8, \sigma_4 = 0.75, \pi_4 = 1/9$

## Appendix C

The following six distributions with $C = 3$ clusters in $d = 2$ or $d = 3$ dimensions are used. In the following, we list the parameters with a brief description along with different distributions.

$\boldsymbol{\mu}$: The location vector.

$\Sigma$: The scatter matrix.

$\nu$: The degrees of freedom.

$\lambda$: The scale parameter.

$\boldsymbol{a}$ and $\boldsymbol{b}$: The parameters of a uniform distribution.

$\boldsymbol{\alpha}$ and $\boldsymbol{\gamma}$: The parameters of multivariate Gamma distribution parameters (see [50] for details).

When $d = 2$, only the values corresponding to the first two dimensions are used. $I_d$ is the identity matrix of $d$ dimensions.

1) Student's $t$ ($\nu = 2$): $\boldsymbol{\mu}_1 = [0\ \ 0\ \ 0], \boldsymbol{\mu}_2 = [20\ \ 15\ \ 10], \boldsymbol{\mu}_3 = [15\ \ -15\ \ 10]$, $\Sigma = I_3$.

2) Cauchy: $\boldsymbol{\mu}_1 = [0\ \ 0\ \ 0], \boldsymbol{\mu}_2 = [20\ \ 15\ \ 10], \boldsymbol{\mu}_3 = [15\ \ -15\ \ 10]$, $\Sigma = I_3$.

3) Uniform: $\boldsymbol{a}_1 = [-5\ \ -5\ \ -5], \boldsymbol{a}_2 = [5\ \ 5\ \ 5], \boldsymbol{a}_3 = [10\ \ 5\ \ 5]$, $\boldsymbol{b}_1 = [6\ \ 6\ \ 6]$, $\boldsymbol{b}_2 = [10\ \ 10\ \ 10]$, $\boldsymbol{b}_3 = [20\ \ 20\ \ 20]$.

4) Gamma: $\boldsymbol{\mu}_1 = [0\ \ 0\ \ 0], \boldsymbol{\mu}_2 = [0\ \ -2\ \ 10], \boldsymbol{\mu}_3 = [10\ \ 10\ \ 10]$, $\boldsymbol{\alpha}_1 = [1\ \ 2\ \ 4\ \ 4]$, $\boldsymbol{\alpha}_2 = [0.5\ \ 1\ \ 2\ \ 2]$, $\boldsymbol{\alpha}_3 = [2\ \ 2\ \ 5\ \ 5]$, $\boldsymbol{\gamma}_1 = [1\ \ 1\ \ 1], \boldsymbol{\gamma}_2 = [1\ \ 1\ \ 1], \boldsymbol{\gamma}_3 = [1\ \ 1\ \ 1]$.

5) Gaussian with uniform noise: $\boldsymbol{\mu}_1 = [0\ \ 0\ \ 0], \boldsymbol{\mu}_2 = [2\ \ 3\ \ 2], \boldsymbol{\mu}_3 = [5\ \ -2\ \ 10]$, $\Sigma_1 = I_3$, $\Sigma_2 = \Sigma_3 = 2 * I_3$, $\boldsymbol{a} = [-30\ \ -30\ \ -30]$, $\boldsymbol{b} = [40\ \ 40\ \ 40]$, $n_{noise} = d * 50$.

6) Gaussian-Laplace mixture: $\boldsymbol{\mu}_{G1} = \boldsymbol{\mu}_{L1} = [0\ \ 0\ \ 0]$, $\boldsymbol{\mu}_{G2} = \boldsymbol{\mu}_{L2} = [15\ \ -5\ \ 10]$, $\boldsymbol{\mu}_{G3} = \boldsymbol{\mu}_{L3} = [5\ \ -10\ \ 15]$, $\Sigma = 5 * I_3$, $\lambda = 0.1$, $\pi = 0.5$.

The three clusters have sample size $n_1 = d * 100, n_2 = d * 100, n_3 = d * 50$ (small) and $n_1 = 10 * d * 100$, $n_2 = 10 * d * 100$, $n_3 = 10 * d * 50$ (large). Because the integrals cannot be solved analytically, we used importance sampling from the fitted GMM with 100,000 samples. For the GMM fitting, we used the following parameters: number of repetitions = 10; maximum number of iterations = 1000; number of components = $1, \cdots, 25$; starting points: randomly selected observations. $10^{-6}$ was added to the diagonal of the covariance matrices to ensure that they are positive-definite.